\documentclass[onecolumn]{robbyant}           %

\usepackage{makecell}
\usepackage{wrapfig}
\usepackage{subcaption}
\usepackage{tikz}
\usetikzlibrary{arrows.meta,positioning,calc,fit,backgrounds}
\usepackage{pgf-pie}
\usepackage{pgfplots}
\usepackage{arydshln}
\pgfplotsset{compat=1.18}
\usepackage{amsthm}
\newtheorem{finding}{Finding}
\definecolor{pie1}{HTML}{5B8FF9}
\definecolor{pie2}{HTML}{5AD8A6}
\definecolor{pie3}{HTML}{F6BD16}
\definecolor{pie4}{HTML}{E86452}
\definecolor{pie5}{HTML}{6DC8EC}
\definecolor{pie6}{HTML}{945FB9}
\definecolor{pie7}{HTML}{FF9845}
\definecolor{pie8}{HTML}{1E9493}
\definecolor{pie9}{HTML}{FF99C3}

\metadata[Website]{\url{https://technology.robbyant.com/lingbot-vision}}
\metadata[Github]{\url{https://github.com/robbyant/lingbot-vision}}
\metadata[Checkpoints]{\url{https://huggingface.co/collections/robbyant/lingbot-vision}}
\newcommand{\methodname}{LingBot-Vision}
\newcommand{\method}{\texttt{\methodname}\xspace}
\newcommand{\taskname}{LingBot-Depth}
\newcommand{\taskold}{\texttt{\taskname}\xspace}
\newcommand{\tasknew}{\texttt{\taskname~2.0}\xspace}

\title{Vision Pretraining for Dense Spatial Perception}

\author{
\begin{center}
    Zelin Fu$^*$ \quad
    Bin Tan$^*$ \quad
    Changjiang Sun \quad
    Shaohui Liu \quad
    Kecheng Zheng \quad
    \\[2pt]
    Yinghao Xu \quad
    Xing Zhu \quad
    Yujun Shen \quad
    Nan Xue$^\dagger$
    \\[12pt]
    $^*$ Equal contributions \qquad
    $^{\dagger}$Project Lead
\end{center}
}

\begin{document}

\abstract{%

Dense spatial perception is essential for physical intelligence, where visual systems are expected to recover structured, metric, and actionable representations from pixel observations.
Modern visual foundation models tend to prioritize semantic invariance, often at the expense of detailed spatial understanding.
In this work, we study vision pretraining through a \textbf{\textit{boundary-centric}} lens, motivated by the premise that boundaries and shape discontinuities offer essential cues for perceiving geometric properties.
Concretely, we propose \textbf{\textit{masked boundary modeling}}, a self-supervised paradigm that dynamically learns sub-pixel boundary representations and subsequently leverages the discovered boundary-bearing tokens as masked targets to facilitate dense visual token learning.
By scaling this framework, we develop \method and demonstrate its efficacy across a diverse set of downstream vision tasks with DINOv3 as a strong baseline.
Remarkably, \method drives the progression from \taskold to \tasknew for depth completion, and thereby yields enhanced depth estimation, a key pillar for embodied artificial intelligence.
Our findings reveal that boundary modeling goes beyond simple line segments and instead serves as a scalable pretraining principle for learning spatially structured visual representations.
}

\maketitle
\justifying

\begin{figure}[!t]
\centering
\begingroup
\resizebox{\linewidth}{!}{\input{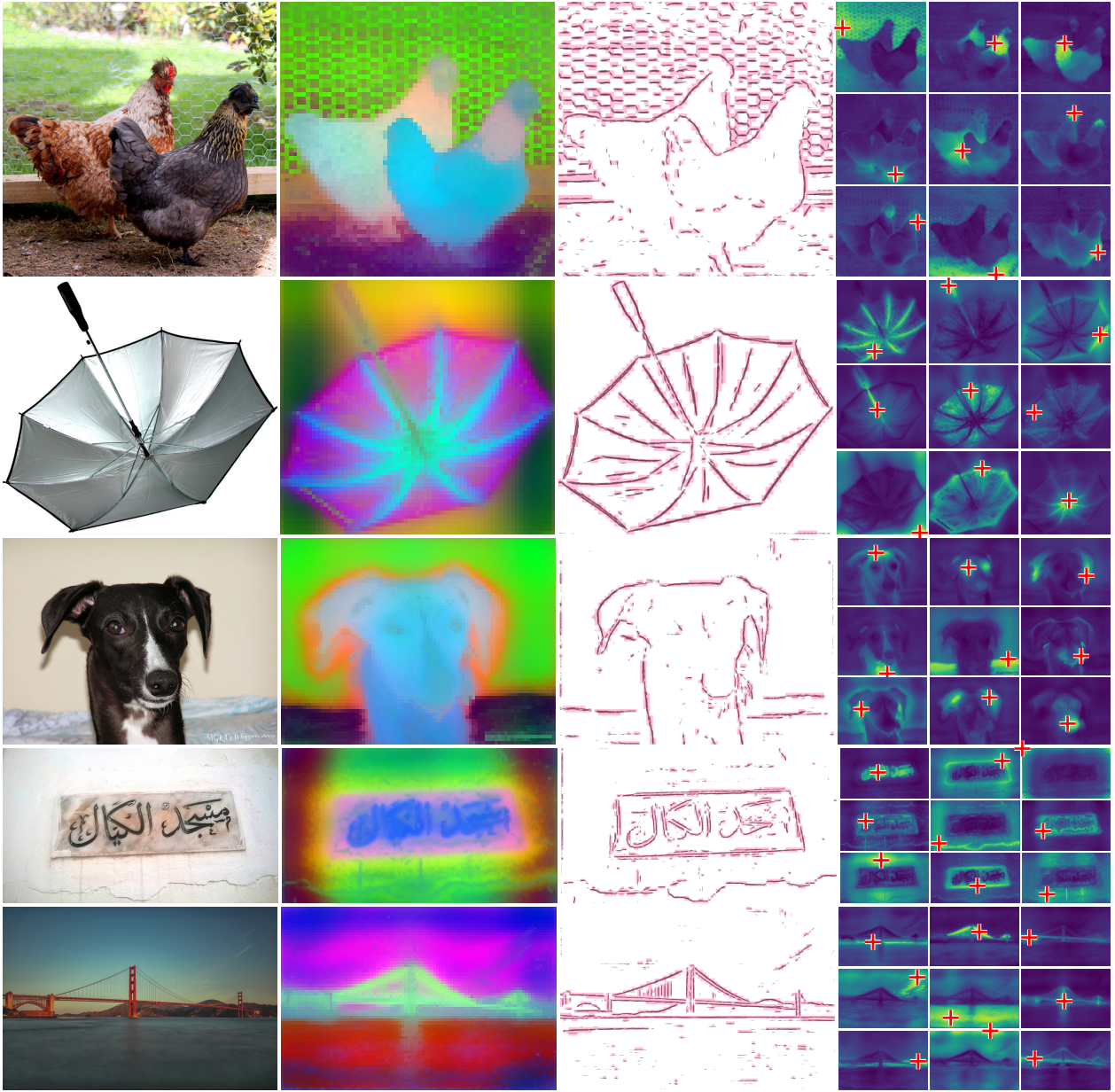}}
\endgroup
\caption{\textbf{\method{} learns dense representations via boundary-centric masked modeling.}
Each row, from left to right: the input image; the PCA projection of the frozen teacher's patch tokens; boundary tokens (pink) obtained by a-contrario validation of dense line proposals decoded from the model's own boundary-field prediction, overlaid on the accumulated response of the validated proposals; and cosine-similarity maps between nine boundary-token queries (red crosses, selected by farthest-point sampling in feature space) and all patch tokens.
The learned representations carry both semantic grouping and geometric structures.
Input images are at 1024px for the short side.
}
\label{fig:teaser}
\end{figure}

\section{Introduction}
Visual intelligence is not only about recognizing \emph{what} is in an image, but also about recovering the dense spatial structure that makes the physical world measurable, navigable, and actionable. Across segmentation, depth, motion, and scene layout, this structure is repeatedly revealed by shapes and boundaries: object masks delineate entities, depth discontinuities expose geometry, motion silhouettes separate dynamic regions, and occlusion contours organize visible surfaces. 
Yet many modern visual foundation models are driven by objectives that favor semantic invariance, cross-modal alignment, or appearance reconstruction, and remain comparatively weak at the fine-grained spatial understanding that downstream physical intelligence depends on.

A central reason is how shapes and boundaries are treated. They are usually regarded as \emph{outputs} of perception, recovered by task-specific heads under dedicated supervision. Such an output-centric view ties them to annotations that are expensive, ambiguous, and often unavailable, making them difficult to exploit in large-scale pretraining. As a result, foundation models rarely use shapes and boundaries as \emph{native} learning signals, relying instead on view-invariant self-distillation~\cite{dino,oquab2023dinov2,simeoni2025dinov3}, cross-modal alignment~\cite{clip,siglip}, or masked image reconstruction~\cite{mae}. We argue that this is a missed opportunity: boundaries and shape discontinuities are not merely outputs to be predicted, but fundamental organizing signals for learning dense representations.

In this work, we therefore study how randomly initialized Vision Transformers can learn shape- and boundary-aware representations from raw images alone, without human annotations, external edge detectors, or pretrained backbones. Our study builds on dense, over-parameterized field representations of boundary geometry~\cite{hawp,hat}, which encode vectorized boundaries through pixel-wise, distance-transform-like fields. Such representations have been extensively used in supervised line-segment and wireframe detection, where they turn sparse geometric structures into dense learning signals. While they have rarely been considered for self-supervised representation learning, we find that their dense and over-parameterized nature makes them uniquely well suited to our goal: \emph{boundary geometry can be bootstrapped from raw images during pretraining}, with usable boundary fields emerging even from an untrained model.

We adopt a self-distillation formulation, but extend its teacher--student mechanism from semantic token learning to boundary-aware dense token learning. Starting from a randomly initialized network, the teacher generates online targets for both global visual semantics and sub-token boundary fields, while the student learns to recover them from masked views; through EMA updates, the teacher's boundary perception co-evolves with the student. This creates a fundamental tension: semantic self-distillation may suppress fine-grained boundary signals, while boundary modeling may in turn disrupt the invariance needed for strong semantic representations. The core of our method is a design that turns this tension into cooperation.

\paragraph{Boundary-Forcing Masked Modeling.}
Our key idea is to let the teacher's own boundary predictions decide \emph{what the student must reconstruct}. From the teacher's online boundary fields we identify the boundary-bearing tokens and \emph{force} them into the student's masked set, so the student must recover boundary geometry purely from surrounding context. This turns boundaries, the most information-dense and least redundant regions of an image, into the hardest and most informative prediction targets. Crucially, the masked tokens are then \emph{routed by geometry}: boundary tokens are supervised by the discovered boundary fields, while the remaining masked tokens follow the standard semantic self-distillation objective. Because a semantic target is inherently ambiguous exactly where two regions meet, this routing supplies a well-posed objective precisely where conventional masked modeling is weakest. It is what allows boundary-aware and semantic representations to \emph{co-emerge} rather than compete, resolving the tension above.

\paragraph{Categorical Reparameterization of Boundary Fields.}
A second ingredient makes dense boundary self-distillation stable. Directly regressing continuous boundary fields in a teacher--student loop is prone to collapse. We instead \emph{reparameterize} each boundary field into a categorical distribution over discretized bins, recasting boundary prediction as a per-pixel classification problem. This brings two benefits. First, the boundary objective becomes compatible with the same centering and sharpening mechanisms that stabilize semantic self-distillation, so boundary learning inherits the scalability of modern SSL~\cite{oquab2023dinov2}. Second, it connects to classical a-contrario detection theory: under the no-structure null hypothesis, boundary orientations are uniformly distributed, so deviation from uniformity provides a principled, parameter-free measure of how strongly a token supports a real boundary, a validation signal obtained for free.

\paragraph{Scaling and Results.}
We instantiate masked boundary modeling at scale as \method, a one-billion-parameter Vision Transformer trained purely with self-supervision. Across dense spatial tasks, \method matches or surpasses visual foundation models up to seven times larger: it attains the best NYU-Depth~v2 accuracy of all compared models, including the 7B DINOv3, stays on par with distilled dense specialists on semantic and video object segmentation, and yields boundary tokens stable enough to be tracked through video by plain cosine similarity of the frozen features. These advantages survive distillation: the ViT-L, ViT-B and ViT-S students lead dense prediction in their size classes, and the 0.3B student matches the NYU-Depth~v2 accuracy of the 7B DINOv3 with roughly $23\times$ fewer parameters. Finally, we advance our depth completion system from \taskold to \tasknew by changing nothing in its recipe but the encoder initialization and the scale of the training data, obtaining leading performance on 14 depth completion benchmarks; notably, the advantage of the \method encoder widens as the downstream training data grows, so better pretraining compounds instead of washing out. Together, these results show that boundary modeling, long confined to standalone contour or line-segment detection, can serve as a general and scalable pretraining principle for learning spatially structured visual representations.

\paragraph{Contributions.} 
\begin{itemize}
    \item We advocate a \emph{boundary-centric} view of self-supervised pretraining, in which boundaries are native learning signals rather than downstream outputs, and show they can be bootstrapped from raw images without human labels, external edge detectors, or pretrained backbones.
    \item We propose \emph{masked boundary modeling}, a self-distillation framework that forces teacher-discovered boundary tokens into the student's masked set and routes masked tokens by geometry, turning the conflict between semantic abstraction and geometric sensitivity into cooperation.
    \item We introduce a \emph{categorical reparameterization} of boundary fields that stabilizes dense self-distillation and yields parameter-free boundary validation at no extra cost.
    \item We scale the framework to \method, a 1B Vision Transformer that rivals or surpasses foundation models up to $7\times$ larger on dense spatial perception, distill it into a strong ViT-L/B/S family, and deliver \tasknew with leading results on 14 depth completion benchmarks.
\end{itemize}

\section{Related Work}
In this section, we briefly revisit the development of representation learning for vision data with a focus on self-supervised pretraining, and then discuss boundary modeling in the literature and its connection to self-supervised vision pretraining.

\subsection{Self-supervised Visual Representation Learning}
Learning meaningful and versatile representations from visual data has long been a central topic in computer vision and machine learning.  Self-supervised learning (SSL), one of the most promising directions, aims to learn representations from the data itself, without labels, and the last decade has witnessed its great success through contrastive learning~\cite{moco,simclr}, masked image modeling~\cite{mae,ibot,pixio}, self-distillation~\cite{dino,oquab2023dinov2,simeoni2025dinov3,fan2025webdino,venkataramanan2025franca,tipsv2}, and most recently the joint-embedding predictive architectures (JEPA)~\cite{ijepa,vjepa,vjepa2p1}. Over the years, the advances in visual SSL have been shifting from
global to dense representations, moving toward larger-scale training,
and seeking more principled ways to avoid collapse at scale.

\paragraph{Pretraining using Self-Distillation.}
The self-distillation lineage currently dominates. DINO~\cite{dino} trains a student to match the categorical token distributions of an EMA teacher and, notably, was among the first to show object layouts and boundaries emerging in attention maps without any supervision.  iBOT~\cite{ibot} moved the same agreement objective down to masked tokens, DINOv2~\cite{oquab2023dinov2} scaled it on curated data into a distilled model family, Web-SSL~\cite{fan2025webdino} showed that language-free pretraining keeps scaling on billions of web images, and Franca~\cite{venkataramanan2025franca} reworked the clustering objective itself. The weakness of the lineage is equally consistent: because the objective rewards agreement rather than spatial structure, dense feature quality is fragile at scale, and DINOv3~\cite{simeoni2025dinov3} had to introduce Gram anchoring precisely to stop dense feature maps from degrading over long training schedules. Boundaries and layout emerge in these models as a byproduct, and keeping them intact requires ever more corrective machinery.

\paragraph{Masked Image Modeling.}
The two other families occupy opposite ends of a spatial-semantic trade-off. Masked image modeling~\cite{beit,mae,pixio} reconstructs missing content against pixel-level targets, which makes it spatially faithful by construction but comparatively weak as a frozen semantic representation.  Language supervision, from CLIP~\cite{clip} to SigLIP~2~\cite{tschannen2025siglip}, is the mirror image: text describes whole images, so the learned features excel at recognition while underperforming on dense prediction, and even finer patch-text alignment~\cite{tipsv2} recovers only part of the gap. Neither extreme delivers representations that are simultaneously semantic and spatially
precise.

\paragraph{Predictive Architectures and World Models.}
JEPA methods~\cite{ijepa,vjepa,assran2025v} predict in latent space to avoid modeling pixel-level nuisance, and LeJEPA~\cite{lejepa} recently grounded the family in a theoretically justified objective. Their ambition raises the stakes for spatial quality: frozen encoders are becoming the perception substrate of world models and robot policies~\cite{dinowm,lewm}, where an error in geometry is no longer a lost benchmark point but a wrong action.
 
Across all of these families the same pattern recurs: dense spatial quality is obtained indirectly and then protected by retrofits, whether post-hoc feature upsamplers~\cite{jafar}, corrective regularization such as Gram anchoring, or agglomerative distillation from several teachers~\cite{ranzinger2024radio,bolya2025perception}. \method takes the trend of the field to its conclusion and makes dense spatial structure the objective of pretraining itself.

\subsection{Boundary Structure Modeling of Images}
Boundary structures, edges and contours are among the oldest
formalizations of image content, capturing the spatial intensity
changes of images, and have been studied and applied since the
beginning of computer vision, well known through the operators of
Sobel~\cite{sobel} and Canny~\cite{canny} and their statistical and structured successors~\cite{lsd,struct-edges}. The modern practice of boundary modeling rests on learning with deep neural networks, ranging from supervised to unsupervised and self-supervised learning~\cite{hed,afm,hawp,hat,wireframe,lcnn,deeplsd,scalelsd}.  Within this body of work, our pretraining inherits two ideas directly. The first is statistical: the a-contrario theory~\cite{desolneux} declares a structure meaningful only if it would be unlikely under a structureless null model, which gave LSD~\cite{lsd} its parameter-free control of false detections. The second is representational: instead of classifying pixels as boundary or not~\cite{hed}, the attraction-field line~\cite{afm,hawp,hat} encodes each line segment redundantly into a dense field in which every nearby pixel stores enough geometry to reconstruct the entire segment, turning sparse detection into dense prediction~\cite{wireframe,lcnn}; subsequent work carried this representation to self-supervised training at scale~\cite{deeplsd,scalelsd}. Throughout this literature, the boundary is the product and the field is the means. Our work inverts the relationship: the attraction field, made categorical, becomes the medium of self-supervised pretraining; its uniform distribution doubles as the a-contrario null hypothesis; and detection-quality boundaries fall out as a byproduct of representation learning rather than being its goal.

\section{Boundary-centric Masked Representation Modeling}
\label{sec:method}
\begin{figure}[t]
\centering
\begin{subfigure}[t]{0.24\linewidth}
  \includegraphics[width=\linewidth]{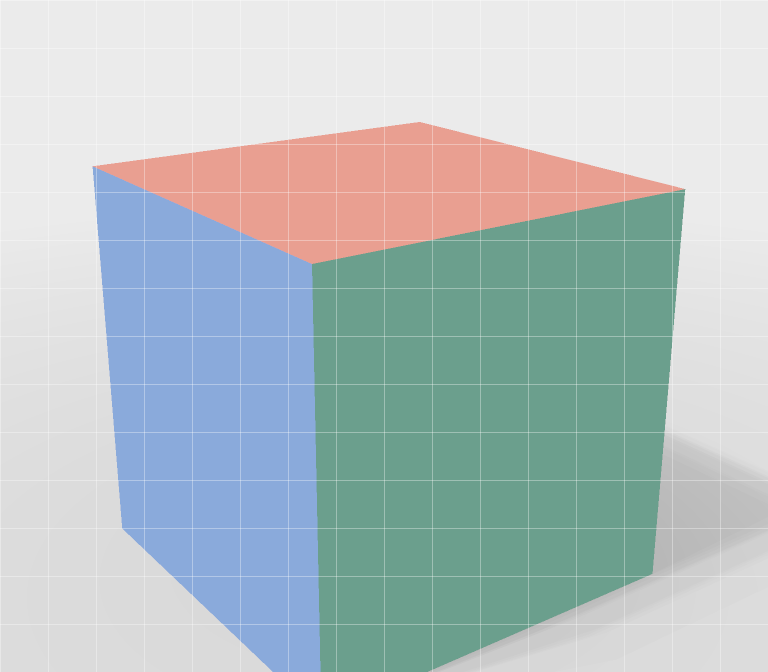}
  \caption{input image}
\end{subfigure}\hfill
\begin{subfigure}[t]{0.24\linewidth}
  \includegraphics[width=\linewidth]{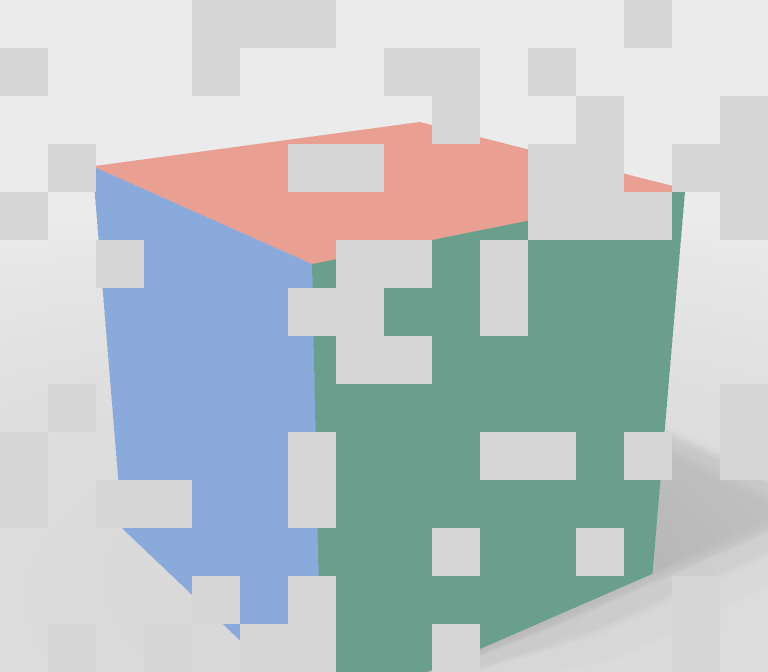}
  \caption{random masking}
\end{subfigure}\hfill
\begin{subfigure}[t]{0.24\linewidth}
  \includegraphics[width=\linewidth]{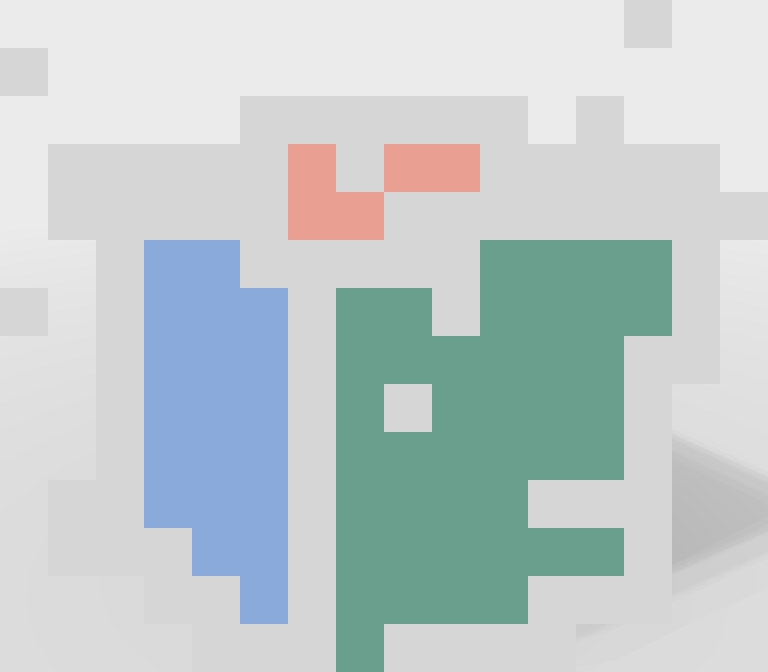}
  \caption{boundary-forcing (ours)}
\end{subfigure}\hfill
\begin{subfigure}[t]{0.24\linewidth}
  \includegraphics[width=\linewidth]{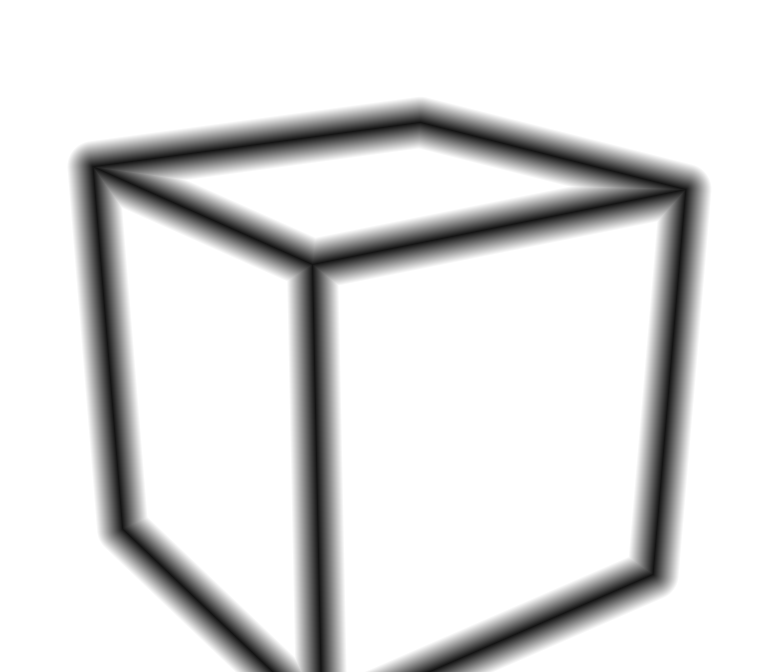}
  \caption{boundary field (target)}
\end{subfigure}
\caption{\textbf{Boundary-centric masked modeling on a toy scene.}
(a) An input image and its patch grid. (b) Random masking is content-agnostic:
most hidden patches are flat and recoverable from context, while the
boundary-bearing patches largely escape the mask. (c) Boundary-forcing masking
adds every boundary-bearing patch to the random mask, so the structure of the
scene is exactly what the student must reconstruct. (d) The boundary field that
supervises boundary tokens, encoded as per-pixel categorical distributions over
discretized distance and orientation bins. During pretraining the boundary
field is predicted online by the teacher itself, without human labels or
external detectors; every masked token follows the semantic self-distillation
objective, and boundary tokens additionally match the categorical boundary
target.}
\label{fig:overview}
\end{figure}

Masked modeling has been a central pretraining paradigm in both natural language processing and computer vision over the past decade. In language, text is a sequence of discrete tokens, so masking a token and predicting it from context is a well-posed problem over a finite vocabulary, the formulation behind masked language models such as BERT (and, in autoregressive form, modern LLMs). Images admit no such natural tokenization: they are continuous, highly redundant arrays of pixels with no discrete units comparable to words. 
Masked modeling in vision therefore first imposes a tokenization, partitioning an $H\times W$ image into a set of non-overlapping patch tokens via a patch embedding, and predicts masked tokens from the rest. Recent masked-modeling frameworks, including MAE~\cite{mae}, I-JEPA and V-JEPA~\cite{ijepa,vjepa}, and the iBOT objective within DINOv2~\cite{oquab2023dinov2} and DINOv3~\cite{simeoni2025dinov3}, all confront the same question of which tokens to mask and predict, and converge on the same answer: {\bf random masking}.

Random masking, however, is content-agnostic: it hides tokens regardless of what they depict, even though tokens differ greatly in how much structure they carry. In this section we present boundary-centric masked modeling, which lets image structure, specifically its boundaries, decide which tokens to mask and what to reconstruct there. We build on the self-distillation paradigm of DINO~\cite{dino} (recapped in Sec.~\ref{sec:prelim:ssl}) and augment it with a second, geometric channel of supervision that the teacher generates for itself. Figure~\ref{fig:overview} illustrates the masking and routing at its core; the subsections that follow develop each component.

\subsection{Self-Distillation Vision Pretraining}
\label{sec:prelim:ssl}
Modern self-supervised pretraining of Vision Transformers (ViTs) follows a teacher--student \emph{self-distillation} paradigm~\citep{dino,oquab2023dinov2,simeoni2025dinov3}. Let $f_\theta$ be the student network and $f_{\bar\theta}$ a teacher of identical architecture whose weights are an exponential moving average (EMA) of the student,
\begin{equation}
    \bar\theta \leftarrow \lambda\,\bar\theta + (1-\lambda)\,\theta,
\end{equation}
with momentum $\lambda$ annealed toward $1$ during training. Given an image $\mathbf{x}$, a set of augmented views is sampled, typically two global crops and several lower-resolution local crops~\citep{dino}. A ViT encodes each view into a class token $\mathbf{z}^{\texttt{cls}}$ and a sequence of patch tokens $\{\mathbf{z}^{\texttt{p}}_i\}_{i=1}^{N}$.

\paragraph{Image-level distillation (DINO).}
A projection head $h$ maps the class token to a distribution over $C$ prototypes. The teacher distribution is centered and sharpened, and the student is trained to match it across views:
\begin{equation}
    \mathbf{p}^{\texttt{t}} = \mathrm{softmax}\!\left(\frac{h_{\bar\theta}(\mathbf{z}^{\texttt{cls}}) - \mathbf{c}}{\tau_t}\right),
    \qquad
    \mathbf{p}^{\texttt{s}} = \mathrm{softmax}\!\left(\frac{h_{\theta}(\mathbf{z}^{\texttt{cls}})}{\tau_s}\right),
\end{equation}
\begin{equation}
    \mathcal{L}_{\texttt{DINO}} = -\,\mathbf{p}^{\texttt{t}}\!\big(\mathbf{x}^{(1)}\big)^{\!\top} \log \mathbf{p}^{\texttt{s}}\!\big(\mathbf{x}^{(2)}\big),
\end{equation}
summed over mismatched view pairs, with a stop-gradient on the teacher. The temperatures satisfy $\tau_t < \tau_s$ (teacher sharpening), and the center $\mathbf{c}$ (an EMA of teacher outputs, or equivalently a Sinkhorn--Knopp normalization~\citep{oquab2023dinov2}) prevents the trivial collapse to a constant distribution.

\paragraph{Patch-level distillation (iBOT).}
To learn dense representations, a subset $\mathcal{M}\subset\{1,\dots,N\}$ of the student's patch tokens is masked and replaced by a shared mask token~\citep{ibot}. With a patch projection head $g$, the student predicts a distribution at each masked location, distilled from the teacher's distribution for the corresponding \emph{unmasked} token:
\begin{equation}
    \mathcal{L}_{\texttt{iBOT}} = -\frac{1}{|\mathcal{M}|}\sum_{i\in\mathcal{M}} \mathbf{q}^{\texttt{t}}_i{}^{\!\top}\log \mathbf{q}^{\texttt{s}}_i,
    \qquad
    \mathbf{q}_i = \mathrm{softmax}\!\big(\frac{g(\mathbf{z}^{\texttt{p}}_i)-\mathbf{c}_{\rm iBOT}}{\tau}\big).
\end{equation}

\paragraph{The mask is chosen at random.}
A defining, and rarely questioned, property of this paradigm is that the masked set $\mathcal{M}$ is sampled \emph{at random}, independently of image content. Concretely, $\mathcal{M}$ is drawn from a block-wise random masking distribution with a target ratio $r$, $\mathcal{M}\sim p_{\mathrm{mask}}(\cdot\,;r)$, so that every patch is equally likely to be hidden regardless of what it depicts. This content-agnostic choice treats all tokens as interchangeable reconstruction targets. Yet the difficulty and informativeness of recovering a token vary enormously across an image: a token in a flat, homogeneous region is largely predictable from its neighbors, whereas a token straddling a boundary carries irreducible structural information.

\subsection{Boundary-Forcing Masked Modeling}
\label{sec:method:forcing}
Sec.~\ref{sec:prelim:ssl} leaves one degree of freedom in masked self-distillation unused: which tokens to hide. Random masking treats all tokens alike, yet boundaries carry structural information that interiors do not. We close this gap by letting the self-distillation discover where boundaries lie and masking accordingly.
Classical edge and line-segment detectors read boundaries out at the pixel level, typically with convolutional decoders. Because a Vision Transformer operates at the token level, we instead attach a boundary attribute to each token: a token is a boundary token if a line or curve predicted by the teacher passes through its $P\times P$ patch (here $P{=}16$). How those lines are predicted and validated is the subject of Sec.~\ref{sec:method:bootstrap}.

\paragraph{Boundary-forcing mask and geometry routing.}
Concretely, the teacher's predicted boundaries, produced online without labels (Sec.~\ref{sec:method:bootstrap}), are rasterized to a pixel map. A token is a boundary token when a boundary falls within its patch, giving the set
\begin{equation}
  \mathcal{B} = \big\{\, i \in \{1,\dots,N\} : \text{a predicted boundary intersects } \mathrm{patch}(i) \,\big\},
\end{equation}
which we obtain by max-pooling the boundary map onto the token grid. We then \emph{force} $\mathcal{B}$ into the student's masked set, on top of the random mask $\mathcal{M}$ of Sec.~\ref{sec:prelim:ssl}:
\begin{equation}
  \mathcal{M}^{+} = \mathcal{M} \cup \mathcal{B}.
\end{equation}
Every masked token is supervised by the semantic iBOT objective $\mathcal{L}_{\texttt{iBOT}}$ as in standard self-distillation; boundary tokens \emph{additionally} receive the categorical boundary objective $\mathcal{L}_{\texttt{bnd}}$ (Sec.~\ref{sec:method:categorical}), and the class token follows $\mathcal{L}_{\texttt{DINO}}$ (Fig.~\ref{fig:overview}). Boundary tokens thus carry a \emph{dual} target: the usual semantic reconstruction, plus an explicit geometric one. Boundaries are the least redundant regions of an image: unlike an interior token, which its neighbors largely determine, a boundary token cannot be recovered by copying context. Forcing boundary tokens into the mask therefore hides exactly the information the student cannot infer for free, and compels it to reconstruct boundary geometry from the surrounding context.

\begin{figure}[!t]
\centering
\includegraphics[width=0.185\linewidth]{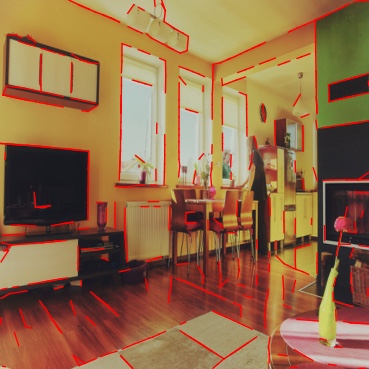}\hfill
\includegraphics[width=0.185\linewidth]{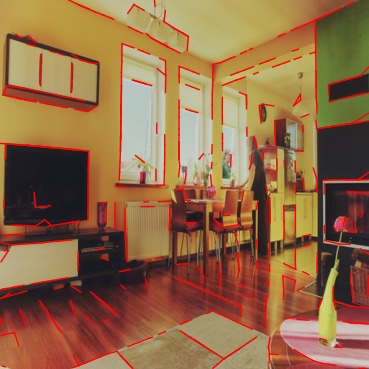}\hfill
\includegraphics[width=0.185\linewidth]{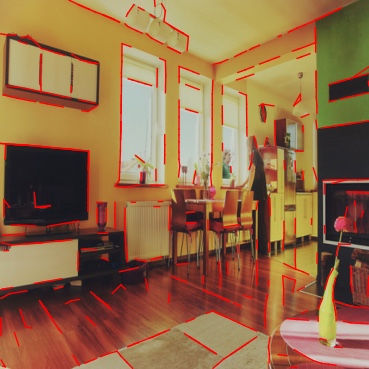}\hfill
\includegraphics[width=0.185\linewidth]{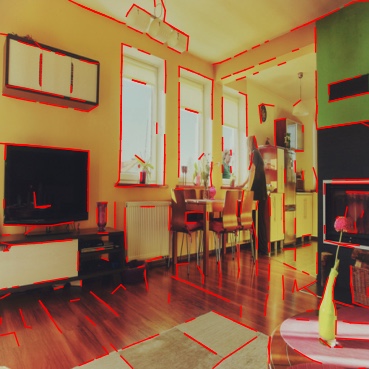}\hfill
\includegraphics[width=0.185\linewidth]{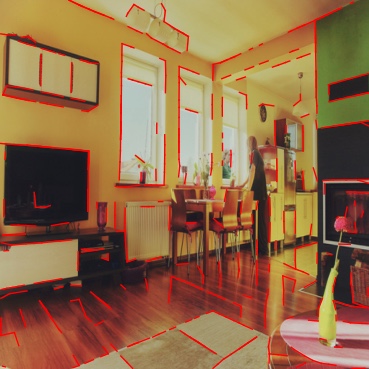}

\vspace{3pt}
\includegraphics[width=0.185\linewidth]{figures/hat_sampling_lsd/000000000724_sample0_lines.jpg}\hfill
\includegraphics[width=0.185\linewidth]{figures/hat_sampling_lsd/000000000724_sample2_lines.jpg}\hfill
\includegraphics[width=0.185\linewidth]{figures/hat_sampling_lsd/000000000724_sample4_lines.jpg}\hfill
\includegraphics[width=0.185\linewidth]{figures/hat_sampling_lsd/000000000724_sample6_lines.jpg}\hfill
\includegraphics[width=0.185\linewidth]{figures/hat_sampling_lsd/000000000724_sample8_lines.jpg}

\vspace{6pt}
\includegraphics[width=0.185\linewidth]{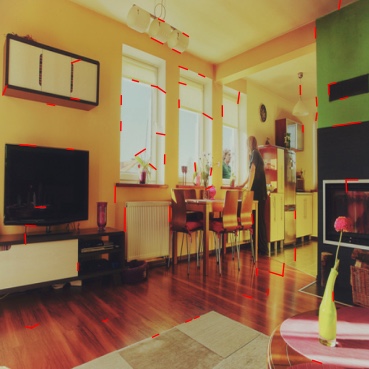}\hfill
\includegraphics[width=0.185\linewidth]{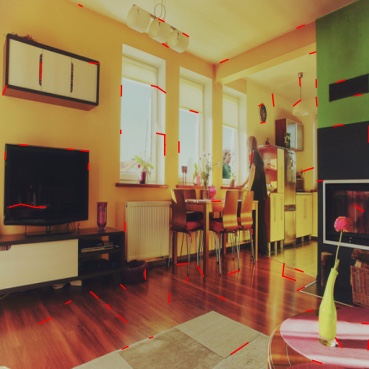}\hfill
\includegraphics[width=0.185\linewidth]{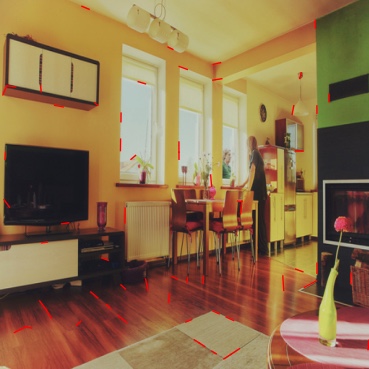}\hfill
\includegraphics[width=0.185\linewidth]{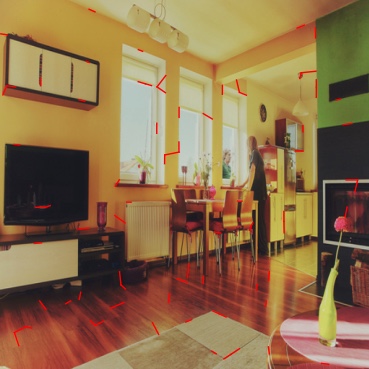}\hfill
\includegraphics[width=0.185\linewidth]{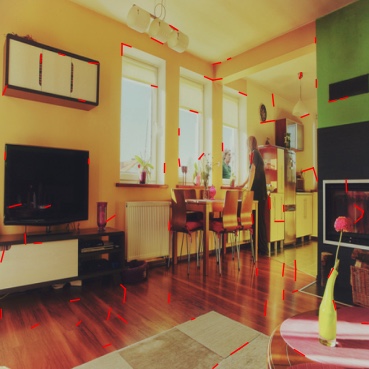}

\vspace{3pt}
\includegraphics[width=0.185\linewidth]{figures/hat_sampling_wolsd/000000000724_sample0_lines.jpg}\hfill
\includegraphics[width=0.185\linewidth]{figures/hat_sampling_wolsd/000000000724_sample2_lines.jpg}\hfill
\includegraphics[width=0.185\linewidth]{figures/hat_sampling_wolsd/000000000724_sample4_lines.jpg}\hfill
\includegraphics[width=0.185\linewidth]{figures/hat_sampling_wolsd/000000000724_sample6_lines.jpg}\hfill
\includegraphics[width=0.185\linewidth]{figures/hat_sampling_wolsd/000000000724_sample8_lines.jpg}
\caption{\textbf{Boundaries emerge from corner points (Finding~\ref{finding:sampling}).}
For two images, we draw five random boundary fields per image, fix the corner points, and decode line segments from each draw (columns). In the bottom two rows, every field channel is sampled \emph{uniformly at random}: the decoded segments still anchor on the corner points, but stay short and fragmented. In the top two rows, only the direction channel is guided by a parameter-free estimate of the image's level-line orientation~\citep{lsd}, rendered into a boundary field, and every draw yields coherent, near-identical segments. In either case the fields carry no learned information. Sampling and decoding details
are in Appendix~\ref{app:sampling}.}
\label{fig:sampling}
\end{figure}

\paragraph{Why the geometric target carries the gain.}
The geometric objective \emph{complements} the semantic one rather than replacing it. A semantic codeword encodes no explicit geometry and is weakest where two or more regions meet, so standard reconstruction under-supervises exactly the boundary tokens we force into the mask; the categorical geometric target supplies the structure it misses. Empirically the two are complementary, not competing: retaining the semantic objective on boundary tokens \emph{alongside} the geometric one improves both classification and dense geometry over the geometric target alone (Sec.~\ref{sec:method:poc}). The geometric target is nonetheless the active ingredient. Our early-stage experiments, which sourced boundaries from an off-the-shelf detector and reconstructed the forced tokens with the semantic objective alone, only matched the random-masking baseline (Sec.~\ref{sec:method:poc}); and such a setup \emph{must} rely on an external detector, because without learning the boundary geometry the teacher has no boundary signal of its own. Learning that geometry is thus what makes boundaries useful to the student: the mask decides \emph{where} the model confronts structure and the categorical objective decides \emph{what} it recovers there, two halves of one mechanism that let boundary-aware and semantic representations co-emerge.

\paragraph{A bootstrapping problem.}
Boundary forcing presupposes that we already know where the boundaries are, yet a network trained from scratch does not: the very tokens we most want to mask are the ones it cannot yet locate. Resolving this chicken-and-egg problem requires clearing two obstacles. First, boundaries must be available from the very first step, before any boundary geometry has been learned. A lightweight corner-point detector makes this possible, thanks to a property of the dense boundary-field representation we build on (Sec.~\ref{sec:method:categorical}):
\begin{finding}[Boundaries emerge from corner points]
\label{finding:sampling}
Given a sparse set of corner points, a boundary field whose values are sampled uniformly at random already decodes into corner-anchored line fragments, and coherent line segments emerge once its direction channel is guided by the parameter-free level-line orientation of the image gradient; without the corner points, neither decoding succeeds.
\end{finding}
Corner points thus anchor usable boundaries from the outset (Fig.~\ref{fig:sampling}), and the fields sharpen as training proceeds (Sec.~\ref{sec:method:bootstrap}); we detail the sampling and decoding procedure in Appendix~\ref{app:sampling}. Second, the boundary target must be learnable without collapsing, which motivates the categorical reparameterization of Sec.~\ref{sec:method:categorical}.

\subsection{Reparameterization of Boundary/Line Fields}
\label{sec:method:categorical}
Finding~\ref{finding:sampling} shows that the holistically-attracted field proposed in~\cite{hat}, hereafter simply the \emph{boundary field}, turns even unlearned, largely random field values into line segments once corner points are given. Thus, if we let the neural network \emph{predict} the field values rather than sample them at random, the decoded segments become image-grounded boundary predictions that improve as the field improves. This section makes the field concrete, explains why the naive way of learning it fails, and derives a categorical reparameterization that makes it learnable.

\begin{figure}[t]
\centering
\begingroup
\input{figures/overview/field_coords.tex}
\begin{subfigure}[t]{0.24\linewidth}\centering
\begin{tikzpicture}[x=0.96\linewidth, y=0.96\linewidth]
  \useasboundingbox (0,0) rectangle (1,0.875);
  \node[anchor=south west, inner sep=0] at (0,0)
    {\includegraphics[width=0.96\linewidth]{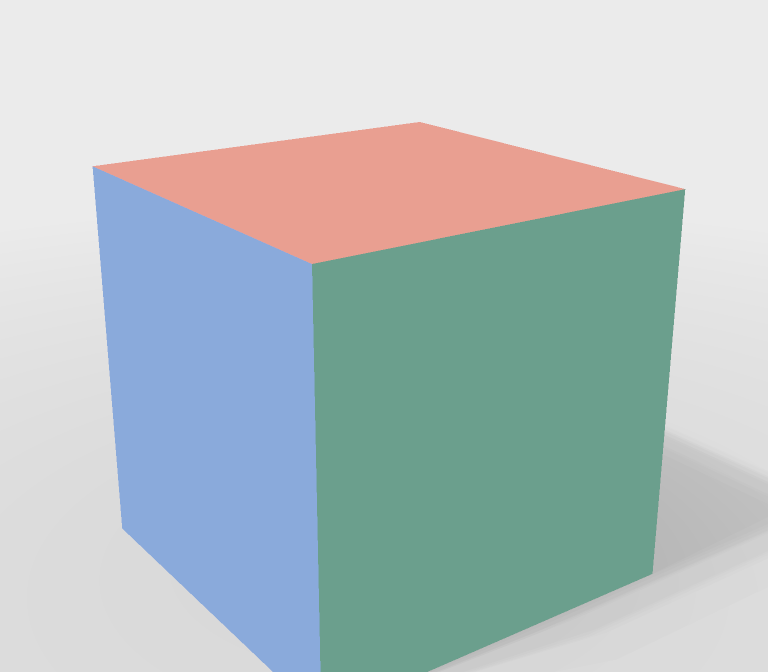}};
  \clip (0,0) rectangle (1,0.875);
  \foreach [count=\i] \ax/\ay/\bx/\by in \fsegs {
    \pgfmathsetmacro{\ci}{int(mod(\i-1,9)+1)}
    \draw[pie\ci, line width=1.0pt, line cap=round] (\ax,\ay) -- (\bx,\by);
  }
\end{tikzpicture}
\caption{boundaries as segments}
\end{subfigure}\hfill
\begin{subfigure}[t]{0.24\linewidth}\centering
\begin{tikzpicture}[x=0.96\linewidth, y=0.96\linewidth]
  \useasboundingbox (0,0) rectangle (1,0.875);
  \node[anchor=south west, inner sep=0] at (0,0.445)
    {\includegraphics[width=0.4914\linewidth]{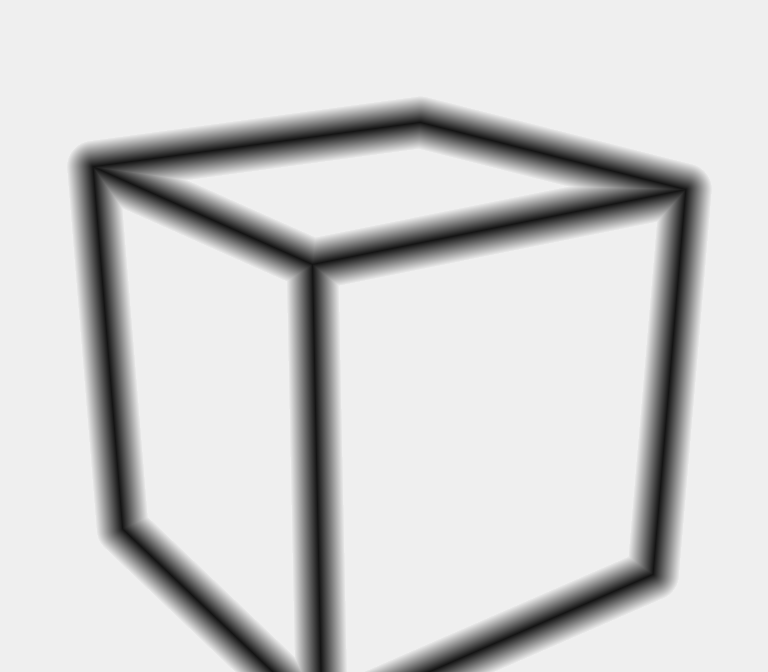}};
  \node[anchor=south west, inner sep=0] at (0.5086,0.445)
    {\includegraphics[width=0.4914\linewidth]{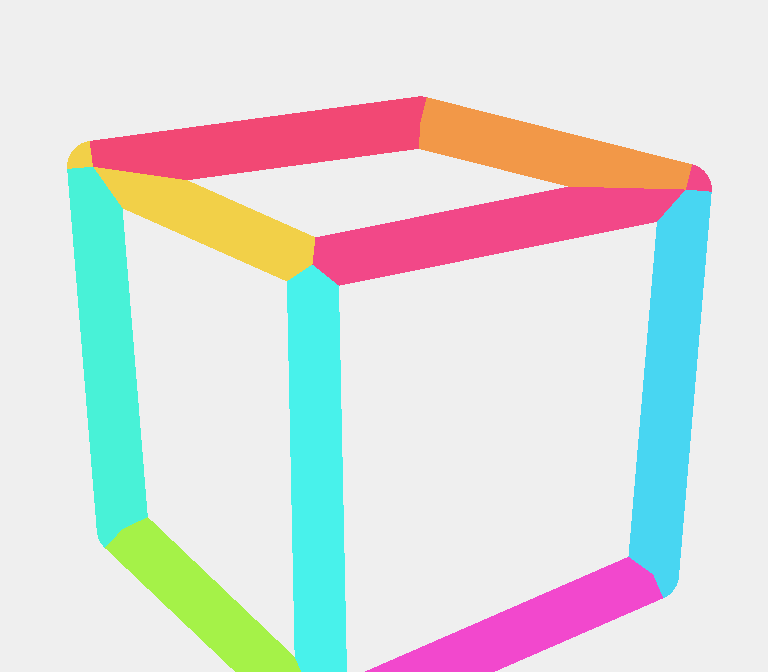}};
  \node[anchor=south west, inner sep=0] at (0,0)
    {\includegraphics[width=0.4914\linewidth]{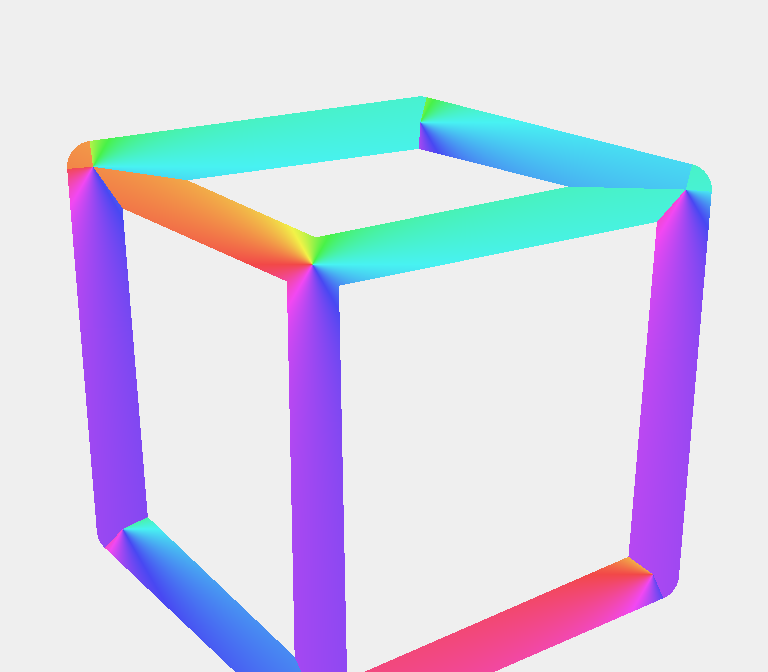}};
  \node[anchor=south west, inner sep=0] at (0.5086,0)
    {\includegraphics[width=0.4914\linewidth]{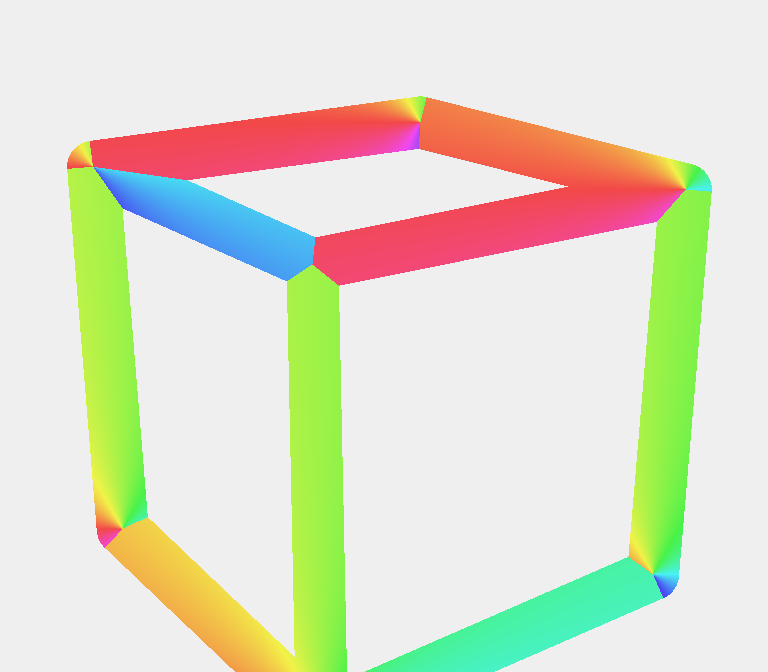}};
  \node[font=\scriptsize, anchor=north west, inner sep=1.5pt]
    at (0.01,0.865) {$d$};
  \node[font=\scriptsize, anchor=north west, inner sep=1.5pt]
    at (0.5186,0.865) {$\theta$};
  \node[font=\scriptsize, anchor=north west, inner sep=1.5pt]
    at (0.01,0.42) {$\phi^{1}$};
  \node[font=\scriptsize, anchor=north west, inner sep=1.5pt]
    at (0.5186,0.42) {$\phi^{2}$};
\end{tikzpicture}
\caption{field channels}
\end{subfigure}\hfill
\begin{subfigure}[t]{0.24\linewidth}\centering
\begin{tikzpicture}[x=0.96\linewidth, y=0.96\linewidth]
  \useasboundingbox (0,0) rectangle (1,0.875);
  \begin{scope}[shift={(0,{0.875-\fsuph})}]
    \node[anchor=south west, inner sep=0] at (0,0)
      {\includegraphics[width=0.96\linewidth]{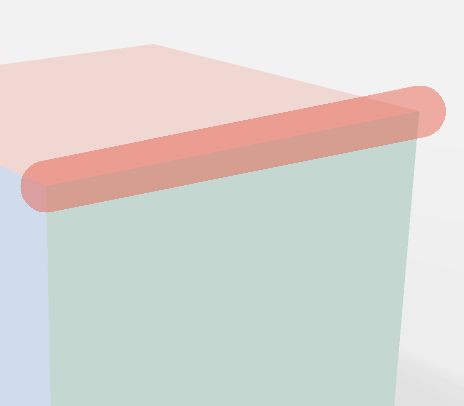}};
    \clip (0,0) rectangle (1,\fsuph);
    \draw[black!70, line width=1.0pt, line cap=round] (\fsupA) -- (\fsupB);
    \fill[black!75] (\fsupA) circle (1.0pt);
    \fill[black!75] (\fsupB) circle (1.0pt);
    \fill[black] (\fq) circle (1.2pt);
    \draw[-{Latex[length=1.7mm]}, black!80, line width=0.7pt] (\fq) -- (\fsupA)
      node[pos=0.65, below=1.5pt, font=\scriptsize, inner sep=1pt] {$\phi^{1}$};
    \draw[-{Latex[length=1.7mm]}, black!80, line width=0.7pt] (\fq) -- (\fsupB)
      node[pos=0.65, below=1.5pt, font=\scriptsize, inner sep=1pt] {$\phi^{2}$};
    \draw[-{Latex[length=1.7mm]}, black!80, line width=0.7pt, densely dashed]
      (\fq) -- (\ffoot) node[pos=0.45, right=1.5pt, font=\scriptsize,
                             inner sep=1pt] {$d$};
    \node[font=\scriptsize, anchor=north, inner sep=2.5pt] at (\fq) {$p$};
  \end{scope}
\end{tikzpicture}
\caption{support region}
\end{subfigure}\hfill
\begin{subfigure}[t]{0.24\linewidth}\centering
\begin{tikzpicture}[x=0.96\linewidth, y=0.96\linewidth]
  \useasboundingbox (0,0) rectangle (1,0.875);
  \node[anchor=south west, inner sep=0] at (0,0)
    {\includegraphics[width=0.96\linewidth]{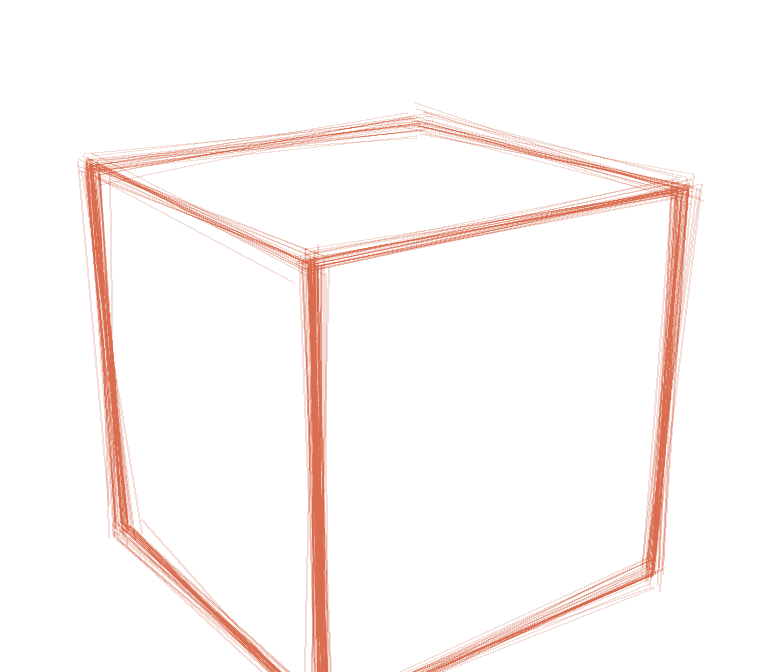}};
  \clip (0,0) rectangle (1,0.875);
  \foreach \ax/\ay/\bx/\by in \fsegs {
    \draw[black!75, line width=0.7pt, line cap=round] (\ax,\ay) -- (\bx,\by);
  }
\end{tikzpicture}
\caption{vote-aggregation decoding}
\end{subfigure}
\endgroup
\caption{\textbf{The boundary field at a glance.}
(a)~Image boundaries are modeled as straight line segments; curved boundaries are chains
of short segments. (b)~The boundary field lifts the sparse segments into a dense map:
every pixel near a segment stores its distance $d$ to the segment and three angles
$(\theta, \phi^{1}, \phi^{2})$ that locate the segment from that pixel; parallel
segments share the same orientation color in $\theta$. (c)~The encoding is deliberately
redundant: any single pixel $p$ in a segment's support region carries enough information
to reconstruct the entire segment. (d)~Conversely, segments are decoded by letting all
support pixels vote and aggregating the votes (orange strokes: individual noisy votes;
dark lines: aggregated segments), which remains reliable even when individual values are
noisy.}
\label{fig:field}
\end{figure}

\paragraph{The boundary field in a nutshell.}
We model image boundaries as a set of straight line segments $L = \{\ell_j\}_{j=1}^{M}$, each given by its two endpoints $\ell_j = (x_1, y_1, x_2, y_2)$; curved boundaries are represented as chains of short segments. The boundary field lifts the sparse set $L$ into a dense map (Fig.~\ref{fig:field}): every pixel $p$ near a segment stores a small attribute vector
\begin{equation}
    \mathbf{a}(p) = \big(d_p,\; \theta_p,\; \phi^{1}_p,\; \phi^{2}_p\big),
\end{equation}
recording its distance $d_p$ to the nearest segment, the direction $\theta_p$ toward it, and two angles $\phi^{1}_p, \phi^{2}_p$ that locate the segment's endpoints as seen from $p$. The parameterization is deliberately redundant: any single pixel in a segment's support region $S_\ell = \{p : d_p \le \tau_d\}$ carries enough information to reconstruct the entire segment, and conversely $L$ is decoded from the field by letting all support pixels vote and aggregating the votes. This many-pixels-one-segment redundancy is what made the decoding of unlearned fields in Finding~\ref{finding:sampling} possible, and what later keeps decoding robust to noisy predictions (Sec.~\ref{sec:method:bootstrap}).

\paragraph{Why line segments rather than edge maps?}
An edge map, which marks photometric discontinuity pixel by pixel, may seem the more general choice. The trouble is that edge pixels can hardly be validated: each is an isolated filter response, accepted or rejected by an arbitrary threshold, with no statistic to test, and texture fires such responses as readily as true structure. A line segment, by contrast, is a single hypothesis supported by its many support pixels, so it can be tested against the no-structure null hypothesis in which orientations are uniformly distributed, and accepted only when chance cannot explain its support (Appendix~\ref{app:nfa}). This difference is decisive in self-distillation, where the teacher's own predictions become the student's targets: unvalidated edges would feed hallucinated structure back into training, whereas validated segments keep the bootstrapped targets clean (Sec.~\ref{sec:method:bootstrap}). Segments moreover carry endpoints and orientation, which is precisely what lets corner points anchor them (Finding~\ref{finding:sampling}).

\paragraph{Why categorical reparameterization?}
The field is a continuous, vector-valued map, and the obvious way to distill it (regressing $\mathbf{a}(p)$ under an $\ell_1$ or $\ell_2$ loss) fails. In a teacher--student loop with an EMA teacher, continuous regression targets drift and collapse; and, unlike the semantic branch, they cannot exploit the centering and sharpening that stabilize self-distillation (Sec.~\ref{sec:prelim:ssl}). We remove this asymmetry by turning boundary prediction into \emph{classification}.

\paragraph{From regression to per-pixel classification.}
We discretize the range of each field channel into $K$ bins and represent a scalar value as a distribution over them. For a boundary position $p$ and channel $c\in\{d,\theta,\phi^{1},\phi^{2}\}$, the (validated) teacher value $a^{c}(p)$ is encoded as a soft categorical label
\begin{equation}
    \bar{y}^{c}_{k}(p) \;\propto\; \exp\!\Big(\!-\,\delta^{c}\big(k,\,a^{c}(p)\big)^{2} \big/ \tau_\ell\Big),
    \qquad k = 1,\dots,K,
\end{equation}
where $\delta^{c}$ is the distance from the center of bin $k$ to the value $a^{c}(p)$ and $\tau_\ell$ a label temperature. For the orientation channel $\theta$, whose range wraps at $2\pi$, the bins are circular and $\delta^{\theta}$ is an arc distance. The label is deliberately \emph{narrow} (a few bins wide): over-smoothing pushes every target toward the uniform distribution and erases the signal. The student's boundary head predicts a distribution $\hat{y}^{c}(p)$ per channel, and the boundary objective is a cross-entropy over the boundary positions $\mathcal{B}$ selected by the forcing mask (Sec.~\ref{sec:method:forcing}),
\begin{equation}
    \mathcal{L}_{\texttt{bnd}} = -\frac{1}{|\mathcal{B}|}\sum_{p\in\mathcal{B}}\;\sum_{c}\; \bar{y}^{c}(p)^{\!\top} \log \hat{y}^{c}(p).
\end{equation}

\paragraph{Inheriting the stability of self-distillation.}
Because boundaries are now categorical, the machinery that keeps semantic self-distillation from collapsing applies unchanged: the teacher distributions are centered (or Sinkhorn--Knopp normalized) and sharpened before they supervise the student, exactly as in Sec.~\ref{sec:prelim:ssl}. This is precisely what the continuous formulation could not do, and it is why boundary learning inherits the stability and scalability of modern SSL instead of requiring bespoke regularizers.

\paragraph{A native null hypothesis, and free validation.}
The categorical form also connects the learning target to classical a-contrario detection theory~\citep{lsd,desolneux}, which validates a candidate segment by asking how unlikely its support would be under a null model containing \emph{no} structure, one in which boundary orientations are uniformly distributed. That null is now literally the uniform distribution over bins. ``No boundary'' is thus a specific, representable prediction, and any departure from uniform is exactly the evidence that the Number-of-False-Alarms (NFA) test measures (Appendix~\ref{app:nfa}). Two consequences follow at no cost: non-boundary regions need no special background class, since their target is simply uniform; and the same distributions that supervise the student can be scored for statistical significance, so boundary validation is \emph{native} to the representation rather than a separate post-hoc step.

\subsection{Online Generation of Boundary Targets}
\label{sec:method:bootstrap}
\begin{figure}[t]
\centering
\begingroup
\input{figures/overview/boot_coords.tex}
\begin{subfigure}[t]{0.24\linewidth}\centering
\begin{tikzpicture}[x=0.96\linewidth, y=0.96\linewidth]
  \node[anchor=south west, inner sep=0] at (0,0)
    {\includegraphics[width=0.96\linewidth]{figures/overview/cube_render.png}};
\end{tikzpicture}
\caption{input image}
\end{subfigure}\hfill
\begin{subfigure}[t]{0.24\linewidth}\centering
\begin{tikzpicture}[x=0.96\linewidth, y=0.96\linewidth]
  \node[anchor=south west, inner sep=0] at (0,0)
    {\includegraphics[width=0.96\linewidth]{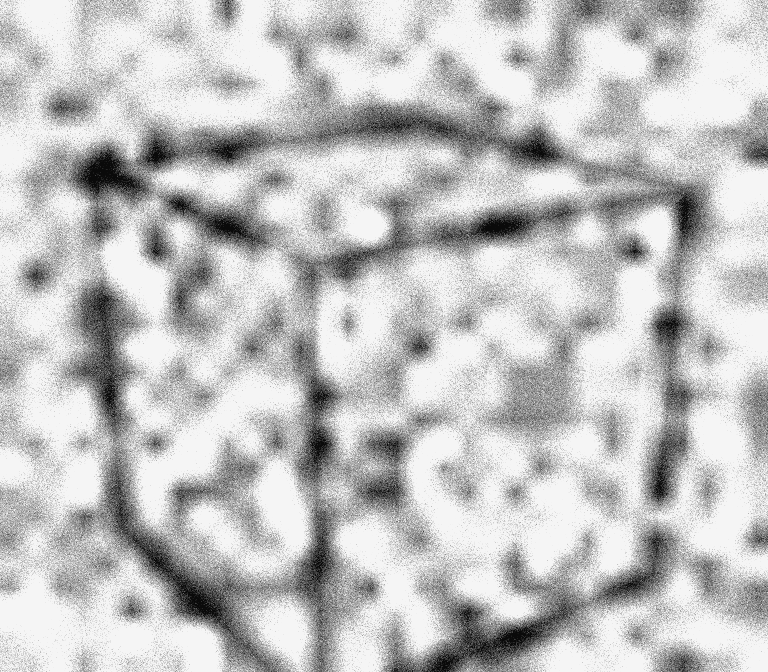}};
  \clip (0,0) rectangle (1,0.875);
  \foreach \cx/\cy in \bootcorners
    {\fill[pie1] (\cx,\cy) circle (1.4pt);
     \draw[white, line width=0.4pt] (\cx,\cy) circle (1.4pt);}
\end{tikzpicture}
\caption{noisy field + corner seeds}
\end{subfigure}\hfill
\begin{subfigure}[t]{0.24\linewidth}\centering
\begin{tikzpicture}[x=0.96\linewidth, y=0.96\linewidth]
  \node[anchor=south west, inner sep=0] at (0,0)
    {\includegraphics[width=0.96\linewidth]{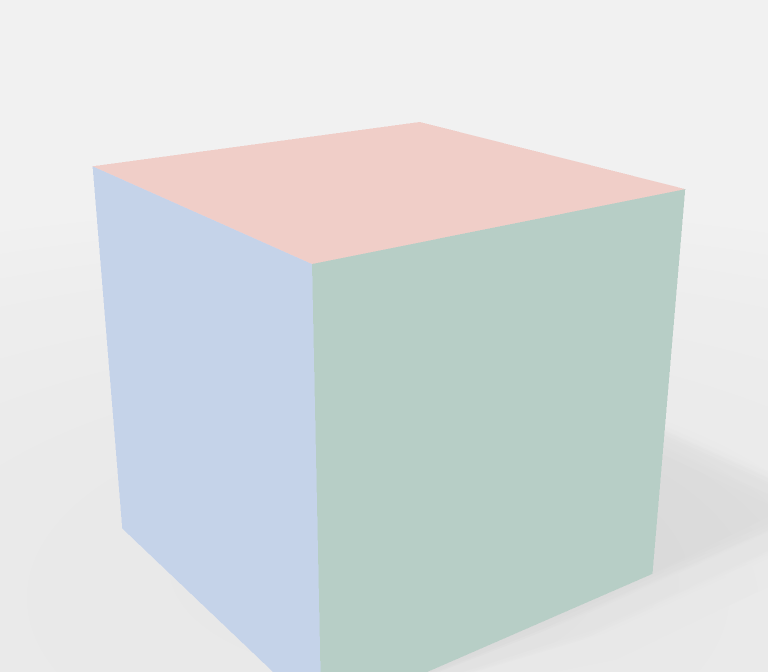}};
  \clip (0,0) rectangle (1,0.875);
  \foreach \ax/\ay/\bx/\by in \bootcands
    {\draw[pie4!85, line width=0.75pt, line cap=round] (\ax,\ay) -- (\bx,\by);}
  \foreach \cx/\cy in \bootcorners
    {\fill[pie1] (\cx,\cy) circle (1.2pt);}
\end{tikzpicture}
\caption{decoded candidates}
\end{subfigure}\hfill
\begin{subfigure}[t]{0.24\linewidth}\centering
\begin{tikzpicture}[x=0.96\linewidth, y=0.96\linewidth]
  \node[anchor=south west, inner sep=0] at (0,0)
    {\includegraphics[width=0.96\linewidth]{figures/overview/cube_boundary.png}};
\end{tikzpicture}
\caption{validated target field}
\end{subfigure}
\endgroup
\caption{\textbf{Online generation of boundary targets (toy example).}
Even when the teacher's boundary field is noisy, as it is early in training (b), the
\emph{holistic} parameterization lets the many weak per-pixel votes in a segment's support
region aggregate into coherent candidate segments anchored at corner points (c), together
with spurious candidates such as face diagonals and background chords. The a-contrario
test then discards unsupported candidates, and the survivors are re-rendered into the
clean target field that supervises the student (d). Boundary targets thus \emph{emerge}
from raw images, with only a frozen single-block corner-point detector fixed
(Sec.~\ref{sec:method:bootstrap}).}
\label{fig:bootstrap}
\end{figure}

As the ViT is initialized from scratch, it carries no information to produce valid boundary targets, yet self-distillation needs a target from the very first step, and we have no boundary annotations to fall back on. Our targets are instead produced \emph{by the teacher itself}, online, and refined as training proceeds. What makes this possible is the redundancy of the boundary field (Sec.~\ref{sec:method:categorical}): because each segment is over-determined by the many pixels in its support region, coherent segments can be read out of the field even when the field is far from accurate.

\paragraph{Boundaries emerge before they are learned.}
Finding~\ref{finding:sampling} already exhibited the extreme case: with corner points fixed, even field values carrying no learned information decode into plausible segments (Fig.~\ref{fig:bootstrap}).
The reason is the voting redundancy above: every support pixel casts a vote for the same underlying segment, and aggregating many weak votes is robust to noise, so the decoded geometry is meaningful long before the field is pixel-accurate. This is the seed that lets boundary perception bootstrap from raw images: the teacher need not be a good boundary predictor to provide a usable signal; it only needs to be better than the student, which the EMA update guarantees.

\paragraph{A token-level head for a dense field.}
The boundary field lives at sub-token resolution, whereas the backbone emits one feature per $P\times P$ token; a lightweight \emph{boundary head} bridges this gap without any convolution. Each patch token is processed independently by a small MLP, and its output is expanded and rearranged into an $r\times r$ tile of positions at stride $s = P/r$ (we use $s{=}2$), so a token literally unfolds into the dense positions it covers. Each position's feature is then $\ell_2$-normalized onto the unit sphere and scored, per field channel, against $K$ learned \emph{bin prototypes}: the bin classifiers are bias-free linear layers with unit-normalized weights, so every logit is a cosine similarity and is bounded by construction. This prototype design mirrors the DINO projection head and complements the centering of Sec.~\ref{sec:method:categorical} in preventing collapse. The teacher and the student share this head design, the teacher's weights being the EMA copy. Finally, continuous field values are read out of the categorical predictions by taking the expectation over bin centers per channel (the circular mean for the orientation channel), so the discrete bins still yield sub-bin.

\paragraph{From a view to a validated target.}
At each iteration, the teacher $f_{\bar\theta}$ turns a global view into a dense target field in four steps (Fig.~\ref{fig:bootstrap}): (i) it predicts a dense boundary field over the view; (ii) it pairs the field with a sparse set of corner points localized by a frozen, single-block Vision Transformer; (iii) combining corner points with the field, it decodes a set of candidate segments $L$ by vote aggregation (Sec.~\ref{sec:method:categorical}); and (iv) it discards segments that fail the a-contrario test (Appendix~\ref{app:nfa}) and re-renders the survivors into a clean, validated target field $\mathbf{a}(\cdot)$. Only this re-rendered field, not the raw prediction, supervises the student, so unsupported structure never becomes a teaching signal. Because the field is regenerated every step from the EMA teacher, the targets co-evolve with the model: as the student sharpens, so does the teacher, and the boundary targets it emits.
Fig.~\ref{fig:anchor_nfa} in Appendix~\ref{app:nfa} ablates the two safeguards of this pipeline on a trained model: reliable corner anchors keep the decoding clean even without validation, and the a-contrario test recovers nearly identical line segments even from weak, field-derived anchors; the decoding degenerates only when both are absent.

\paragraph{No boundary annotations in the loop.}
Throughout, the representation backbone is trained from random initialization, with no human boundary annotation, edge detector, depth sensor, or image-classification pretraining in the loop; the boundary \emph{fields}, the learned target of our method, emerge entirely from raw images. The only fixed component is the \emph{single-block} Vision Transformer that seeds the corner points in step~(ii): it localizes only a sparse set of them, is orders of magnitude smaller than the backbone, and never observes the boundary fields it helps decode. We therefore reserve all representational capacity for learning boundary structure from scratch, rather than importing it from a pretrained model.

\subsection{Training Objective}
\label{sec:method:objective}
The full objective assembles the pieces of Secs.~\ref{sec:prelim:ssl}--\ref{sec:method:categorical}. For each image we sample global and local views; the student processes the global views under the boundary-forced mask $\mathcal{M}^{+}$ (Sec.~\ref{sec:method:forcing}) and is trained to minimize
\begin{equation}
    \mathcal{L} \;=\; \mathcal{L}_{\texttt{DINO}} \;+\; \lambda_{\texttt{i}}\,\mathcal{L}_{\texttt{iBOT}} \;+\; \lambda_{\texttt{b}}\,\mathcal{L}_{\texttt{bnd}} \;+\; \lambda_{\texttt{k}}\,\mathcal{L}_{\texttt{KoLeo}},
    \label{eq:total}
\end{equation}
where $\mathcal{L}_{\texttt{DINO}}$ distills the class token across view pairs, $\mathcal{L}_{\texttt{iBOT}}$ is computed over all masked tokens in $\mathcal{M}^{+}$, and $\mathcal{L}_{\texttt{bnd}}$ is computed over the boundary positions of $\mathcal{B}\subset\mathcal{M}^{+}$, so boundary tokens receive both losses, realizing the dual supervision of Sec.~\ref{sec:method:forcing}. The last term is the KoLeo regularizer~\citep{oquab2023dinov2}, which spreads class-token features within a batch. Loss weights and schedules are deferred to Sec.~\ref{sec:scaling:recipe}.

All teacher quantities, the semantic distributions and the validated boundary targets alike, are produced under stop-gradient, so gradients flow only through the student; after each optimizer step the teacher is updated as the EMA of the student with momentum annealed toward one (Sec.~\ref{sec:prelim:ssl}), and the targets of every branch co-evolve smoothly with the model. Boundary targets are generated for the global views, whose resolution suffices for reliable decoding and validation; local views are supervised by the image-level distillation only. Together with the centering and sharpening applied to both the semantic and the boundary distributions, this is all that is needed to keep training stable: no bespoke regularizer is attached to the boundary branch.

\subsection{Proof of Concept on ImageNet-1K}
\label{sec:method:poc}
Before scaling up (Sec.~\ref{sec:scaling}), we validate each design choice at small scale: ViT-L/16 pretrained on ImageNet-1K~\citep{imagenet}, compared against a DINO+iBOT baseline trained under an identical configuration. We probe the two facets the method is supposed to balance: $k$-NN classification on ImageNet-1K for global semantics, and linear-probe depth estimation on NYU-Depth~v2~\citep{nyuv2} for dense geometry (Table~\ref{tab:poc}).

\begin{table}[t]
\centering
\caption{\textbf{Proof of concept on ImageNet-1K (ViT-L/16).} All variants share the training configuration of the DINO+iBOT baseline.}
\label{tab:poc}
\small
\begin{tabular}{lccc}
\toprule
 & IN-1K $k$-NN $\uparrow$ & \multicolumn{2}{c}{NYUv2 depth (linear)} \\
\cmidrule(lr){3-4}
 & top-1 & $\delta_1 \uparrow$ & RMSE $\downarrow$ \\
\midrule
DINO+iBOT baseline~\citep{dino,ibot}                & 81.6\% & 81.4\% & 0.474 \\
+ categorical boundary target (geometric only)      & 81.8\% & 84.4\% & 0.446 \\
+ dual supervision (iBOT loss on boundary tokens)   & 82.0\% & 84.7\% & 0.443 \\
+ RoPE backbone (final recipe)                      & {\bf 82.4}\% & {\bf 84.9}\% & {\bf 0.440} \\
\midrule
w/ boundary forcing, semantic target only            & 81.4\% & 81.2\% & 0.481 \\
\bottomrule
\end{tabular}
\end{table}

\paragraph{Main comparison.}
The full recipe improves both facets at once: over the matched DINO+iBOT baseline, ImageNet-1K $k$-NN top-1 rises from 81.6\% to 82.4\%, while NYUv2 $\delta_1$ improves from 81.4\% to 84.9\% and RMSE drops from 0.474 to 0.440. Boundary supervision therefore does not trade global semantics for dense geometry: the two co-emerge, as promised in Sec.~\ref{sec:method:forcing}, and the semantic representation itself benefits from confronting structure.

\paragraph{Design ablations.}
The ladder of Table~\ref{tab:poc} attributes the gains. The categorical boundary target is the active ingredient: adding it alone accounts for essentially the entire dense improvement (+3.0 points of $\delta_1$, RMSE from 0.474 to 0.446) at no cost to classification. Dual supervision then adds +0.2 points of $k$-NN and +0.3 points of $\delta_1$ (RMSE 0.446 to 0.443), confirming that the semantic and geometric objectives are complementary rather than competing (Sec.~\ref{sec:method:forcing}). The RoPE backbone contributes a further +0.4 on $k$-NN and +0.2 on $\delta_1$; this is a backbone modernization orthogonal to our method, which we nonetheless carry into the scaled recipe (Sec.~\ref{sec:scaling}). Conversely, forcing boundary tokens into the mask while reconstructing them with the semantic target only performs on par with, indeed slightly below, the baseline ($\delta_1$ 81.2\%, RMSE 0.481), reproducing our early-stage observation under the final pipeline: the mask decides \emph{where}, but the gain comes from \emph{what} is reconstructed there.

\paragraph{What transfers to scale.}
This study freezes the method configuration that Sec.~\ref{sec:scaling} scales unchanged: boundary-forced masking with dual supervision, the categorical boundary target, and the RoPE backbone. Everything that changes from here on, namely the corpus, the model capacity, and the training systems, is external to the method.

\section{LingBot-Vision: Pushing the Boundaries in Vision Pretraining}
\label{sec:scaling}

Based on the promising results in Table~\ref{tab:poc}, we push boundary-centric masked representation learning toward scalable training for large vision foundation models. The method itself is kept exactly as in Sec.~\ref{sec:method}; what scaling changes is everything around it. Following the data curation pipelines of DINOv2 and DINOv3~\cite{oquab2023dinov2,simeoni2025dinov3}, we assemble a corpus of roughly 161M images, selected from 2B raw ones, that combines curated public datasets with retrieval-curated web pools (Sec.~\ref{sec:scaling:data}).
On this corpus we train \method, a ViT-g/16 backbone of roughly 1.1 billion parameters (Sec.~\ref{sec:scaling:model}), supported by the distributed systems that keep the boundary branch a small fraction of the step time (Sec.~\ref{sec:scaling:infra}) and an optimization recipe that keeps training stable at scale (Sec.~\ref{sec:scaling:recipe}). We furthermore find \method to be training-efficient in terms of both data and optimization budget: it consumes a corpus an order of magnitude smaller than DINOv3's and less than one third of its training samples (Secs.~\ref{sec:scaling:data} and~\ref{sec:scaling:recipe}).

\subsection{Training Data Curation}
\label{sec:scaling:data}
Our data curation pipeline follows DINOv2~\cite{oquab2023dinov2}, implemented on our own data sources. The uncurated side is an in-house pool of roughly 2 billion web images; the curated side uses open-source datasets as \emph{seeds} for query-based retrieval from that pool, with image embeddings computed by a pretrained DINOv2 ViT-B encoder~\cite{oquab2023dinov2}. The seeds include ImageNet-1k and ImageNet-21k~\citep{imagenet,imagenet21k}, Google Landmarks v2 (GLDv2)~\citep{gldv2}, and Mapillary-SLS~\citep{mapillary}, together with twenty further datasets spanning fine-grained recognition, instance retrieval, and dense scene understanding (e.g., ADE20K~\citep{ade20k}, NYU-Depth~v2~\citep{nyuv2}, SUN-RGBD~\citep{sunrgbd}, Cityscapes~\citep{cityscapes}, VOC~\citep{voc}), chosen to cover a wide range of downstream tasks.

\paragraph{From 2B raw images to a 161M corpus.}
Each seed contributes in one of two ways, mirroring DINOv2: it is either included \emph{as is}, or it queries the uncurated pool for visually similar images ($k$-NN retrieval for large seeds, cluster-based retrieval for small ones). After retrieval and deduplication, the retrieved portion is materialized as three pools totaling roughly 143M images; the as-is portion comprises ImageNet-21k (13.15M), ImageNet-1k (1.28M), the clean split of GLDv2 (1.58M), and Mapillary-SLS (1.46M). Combined, the corpus holds \textbf{160.75M} images, of which about five sixths are retrieval-curated.
During training, the whole curated corpus is sampled with a single global uniform shuffle, so each image is drawn with equal probability.

\paragraph{Comparison with LVD-142M and LVD-1689M.}
Our corpus is comparable in scale to DINOv2's LVD-142M (142M images curated from a 1.2B pool)~\cite{oquab2023dinov2} and an order of magnitude smaller than DINOv3's LVD-1689M (1,689M images curated from a pool of roughly 17B)~\cite{simeoni2025dinov3}. While we tried our best to curate data, our pipeline is also deliberately simpler than theirs: a single retrieval encoder over a 2B pool, without the concept-rebalancing clustering of LVD-142M or the hierarchical curation of LVD-1689M. That \method nonetheless matches or surpasses models trained on these corpora on dense tasks (Sec.~\ref{sec:exp}) suggests that the gains come from the pretraining objective rather than from a data advantage, and substantiates the training-efficiency claim above.

\subsection{Distributed Training and Efficiency}
\label{sec:scaling:infra}
On the systems side we follow standard practice, sharding the model with FSDP under bf16 mixed precision with memory-efficient attention, and we store the corpus as uncompressed Arrow shards that are memory-mapped for zero-copy random access during loading; none of this is specific to our method. What is specific is the boundary branch, which adds a dense $K$-bin prediction and an online target-generation pipeline on top of standard self-distillation. We describe how both are kept cheap.

\paragraph{Sparse and fused boundary computation.}
The boundary head never runs densely. Only the boundary tokens $\mathcal{B}$, a small fraction of all tokens per batch, are unfolded to sub-token resolution and scored against the bin prototypes, so the head's cost scales with the number of boundary tokens rather than with the image size.
Naively materializing the soft labels and the cross-entropy over four channels and $K$ bins per position would dominate memory traffic; we instead fuse label construction (including the circular arc-distance channel) and the categorical cross-entropy into single custom kernels, and compile the remaining head operations, which removes the multi-gigabyte intermediate tensors entirely.

\paragraph{Online target generation as batched GPU work.}
The teacher-side pipeline of Sec.~\ref{sec:method:bootstrap}, namely field decoding, corner-line pairing, the a-contrario test, and the re-rendering of validated segments, is implemented end-to-end as batched CUDA kernels, with no per-image loops and no host synchronization. Two stages required particular care. First, corner-line pairing computes the distance between every corner point and the two endpoints of every segment proposed by the boundary field; materializing this pairwise distance tensor on the GPU is prohibitively memory-intensive, so we fuse the distance computation with the argmin reduction, keeping memory linear in the number of proposed segments instead of quadratic. Second, the a-contrario test of LSD~\cite{lsd} is a purely sequential CPU design; we reformulate it as a per-candidate reduction over support pixels, so that all candidate segments of a batch are validated in parallel.

An earlier implementation performed segment de-duplication on the CPU and stalled every training step; eliminating all host round-trips was essential for the target generation to overlap cleanly with the rest of the step. In the final recipe, generating validated boundary targets adds only a small fraction of the total step time, and end-to-end training throughput stays close to that of the semantic-only baseline under an identical configuration.

\subsection{Implementation Details}
\label{sec:scaling:recipe}
\label{sec:scaling:model}
Training proceeds in three stages, following the staging of DINOv3~\cite{simeoni2025dinov3}: a long self-distillation pretraining, a Gram-anchoring phase that restores dense-feature quality late in training, and a short high-resolution adaptation. Our budget is deliberately modest. Where DINOv3 spends 1M iterations on low-resolution pretraining, 100k on Gram anchoring, and 30k on multi-resolution adaptation at a global batch size of 4,096, \method spends 300k, 100k, and 100k iterations on the three stages at a global batch size of 3,072; in total, less than one third of the samples DINOv3 processes. To further save compute, our high-resolution adaptation uses a single resolution of $512$ px rather than a multi-resolution schedule.

\paragraph{Backbone and heads.}
\method uses a ViT-g/16 backbone of roughly 1.1B parameters with SwiGLU feed-forward layers, rotary position embeddings computed in fp32, and four register tokens; the proof-of-concept study (Sec.~\ref{sec:method:poc}) uses ViT-L/16. The boundary head follows Sec.~\ref{sec:method:bootstrap}: a three-layer per-token MLP, a head dimension of 512 at output stride $s{=}2$, a learnable tile positional embedding, and $K{=}32$ bins per field channel. The DINO and iBOT projection heads follow DINOv2~\cite{oquab2023dinov2}, with a separate iBOT head. The frozen corner-point detector is a single-block ViT, orders of magnitude smaller than the backbone.

\paragraph{Pretraining.}
We pretrain for 300k iterations with AdamW at a global batch size of 3072, scaling the base learning rate with the square-root rule $\sqrt{\mathrm{bs}/1024}$ and decaying it with a cosine schedule after a linear warmup. Weight decay ramps from 0.04 to 0.2 with a cosine schedule, the teacher temperature ramps from 0.04 to 0.07 in the first 30k iterations, and the EMA teacher momentum is annealed from 0.994 to 1.0 over the course of training.
The objectives of Eq.~\ref{eq:total} are weighted equally ($\lambda_{\texttt{i}} = \lambda_{\texttt{b}} = 1$), with the KoLeo weight $\lambda_{\texttt{k}} = 0.1$, and the boundary branch uses the narrow soft-label temperature of Sec.~\ref{sec:method:categorical}. Two stability practices complete the branch: teacher targets at non-boundary positions are filled with uniform random labels rather than a constant background value, which removes the trivial all-background solution, and teacher-side centering is kept on throughout. Compared with the proof-of-concept configuration, scaling required only mild retuning: a higher initial EMA momentum, a lower weight-decay ceiling, fewer bins ($K$ from 128 to 32, verified to be quality-neutral), and, following DINOv3, plain rather than weight-normalized prototype layers in all heads.
During pretraining, global and local crops are sized $256$ and $112$ px, respectively.

\paragraph{Gram anchoring and high-resolution adaptation.}
Over long schedules, patch-level feature quality is known to degrade even as global metrics keep improving; DINOv3 counteracts this by anchoring the Gram matrix of the student's patch features to that of an earlier teacher snapshot~\cite{simeoni2025dinov3}. We adopt this Gram-anchoring phase for \method, running it for 100k iterations after pretraining. Training then concludes with a 100k-iteration adaptation phase at a single resolution of $512$ px, with a lowered learning rate and a higher EMA momentum, which adapts the backbone (and its rotary embeddings) to high-resolution inputs while the boundary branch remains active.

\paragraph{Model distillation for efficiency.}
After the flagship ViT-g is trained, we distill it into smaller backbones, obtaining a 300M ViT-L, an 86M ViT-B and a 21M ViT-S. Following the distillation practice of DINOv2 and DINOv3~\cite{oquab2023dinov2,simeoni2025dinov3}, the frozen ViT-g replaces the EMA teacher in the same training pipeline, so the smaller students inherit both the semantic and the boundary-aware representations of the flagship. Distillation uses the same iteration budget as the pretraining stage, namely 300k iterations per student; we skip the Gram-anchoring phase, which counteracts a drift that does not arise with a frozen teacher, and apply the high-resolution adaptation at $512$ px for an additional 100k iterations, all at a global batch size of 3,072. The resulting schedule is much shorter than DINOv3's distillation schedule of 1M+250k iterations at a global batch size of 4,096.

\begin{figure}[!tp]
\centering
\setlength{\tabcolsep}{0.5pt}
\renewcommand{\arraystretch}{0}
\begin{tabular}{cccccc}
\includegraphics[width=0.158\linewidth]{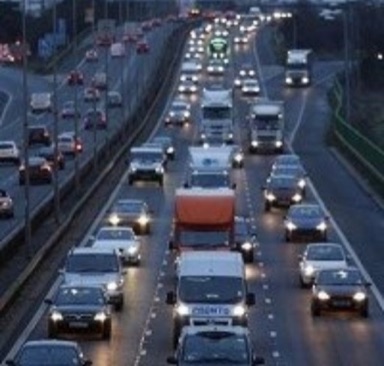} &
\includegraphics[width=0.158\linewidth]{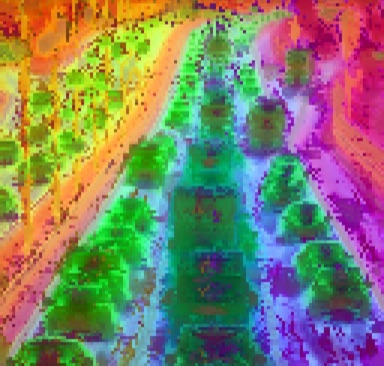} &
\includegraphics[width=0.158\linewidth]{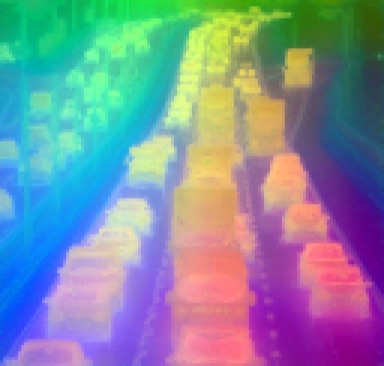} &
\includegraphics[width=0.158\linewidth]{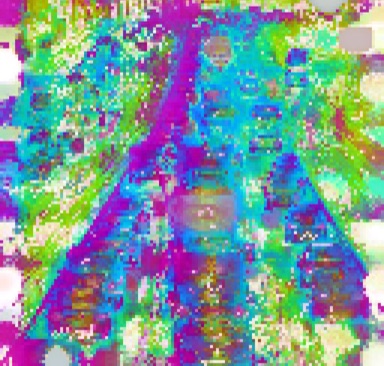} &
\includegraphics[width=0.158\linewidth]{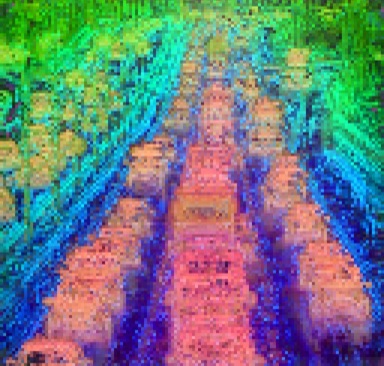} &
\includegraphics[width=0.158\linewidth]{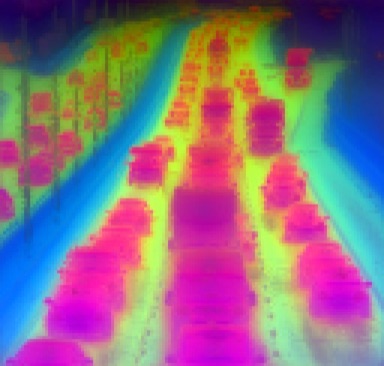} \\
\noalign{\vskip 1pt}
\includegraphics[width=0.158\linewidth]{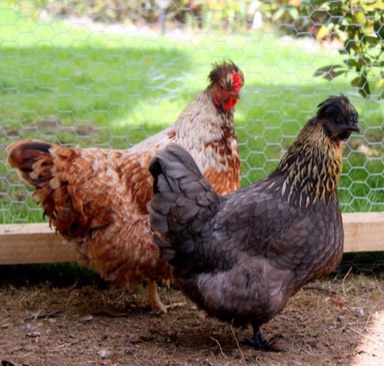} &
\includegraphics[width=0.158\linewidth]{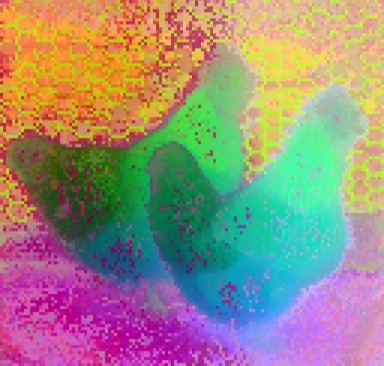} &
\includegraphics[width=0.158\linewidth]{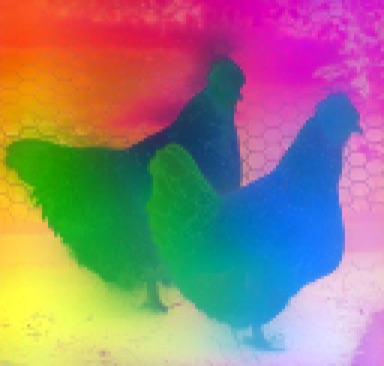} &
\includegraphics[width=0.158\linewidth]{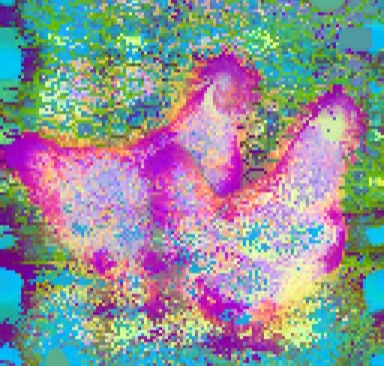} &
\includegraphics[width=0.158\linewidth]{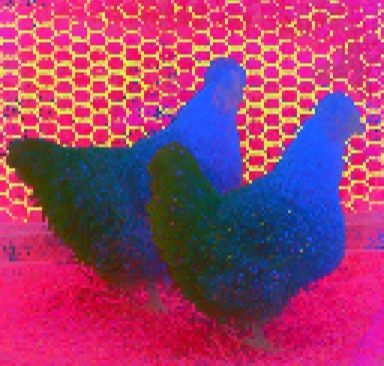} &
\includegraphics[width=0.158\linewidth]{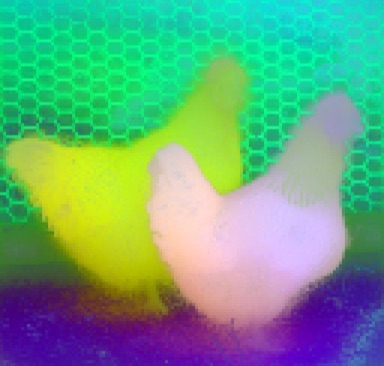} \\
\noalign{\vskip 1pt}
\includegraphics[width=0.158\linewidth]{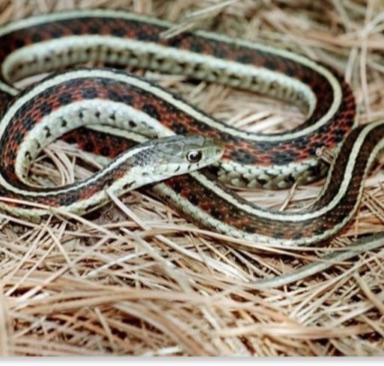} &
\includegraphics[width=0.158\linewidth]{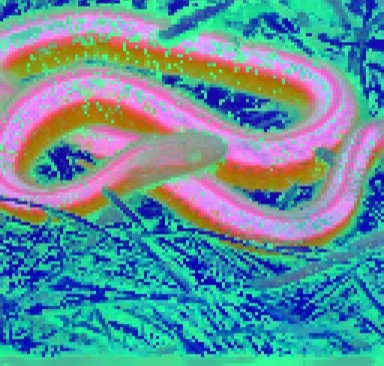} &
\includegraphics[width=0.158\linewidth]{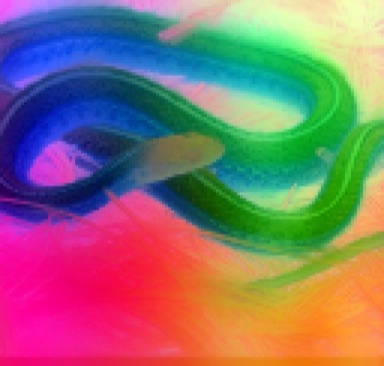} &
\includegraphics[width=0.158\linewidth]{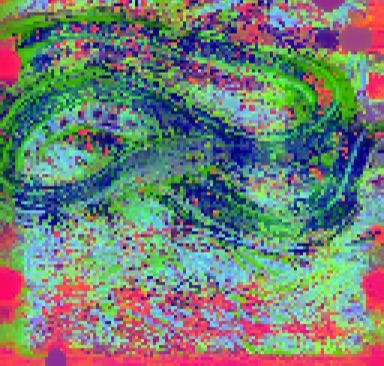} &
\includegraphics[width=0.158\linewidth]{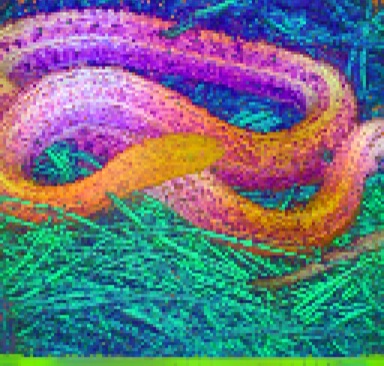} &
\includegraphics[width=0.158\linewidth]{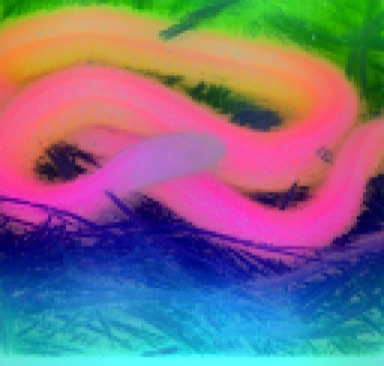} \\
\noalign{\vskip 1pt}
\includegraphics[width=0.158\linewidth]{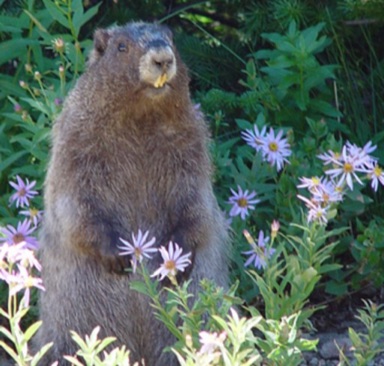} &
\includegraphics[width=0.158\linewidth]{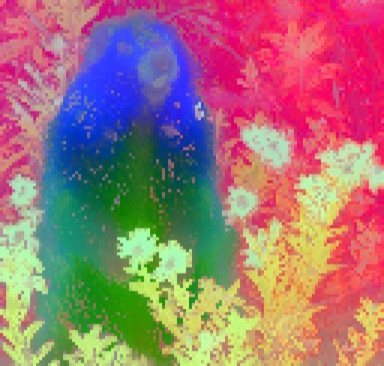} &
\includegraphics[width=0.158\linewidth]{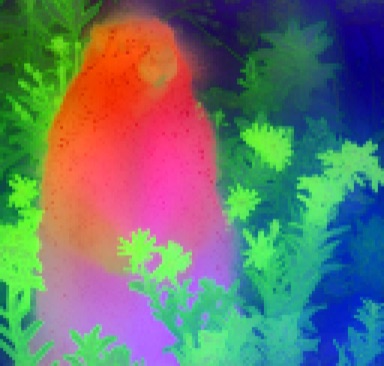} &
\includegraphics[width=0.158\linewidth]{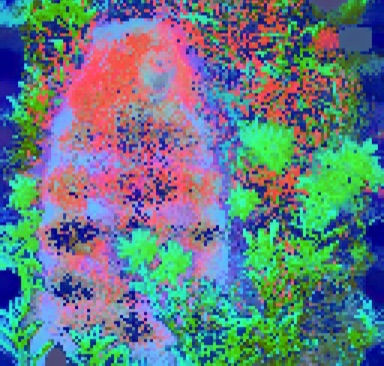} &
\includegraphics[width=0.158\linewidth]{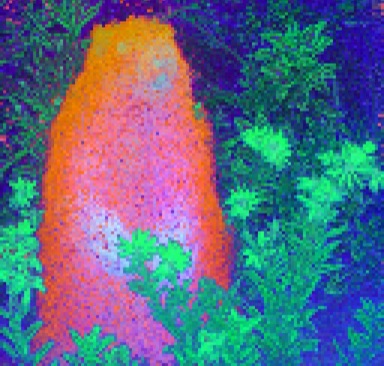} &
\includegraphics[width=0.158\linewidth]{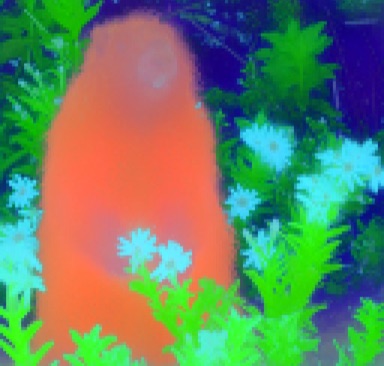} \\
\noalign{\vskip 1pt}
\includegraphics[width=0.158\linewidth]{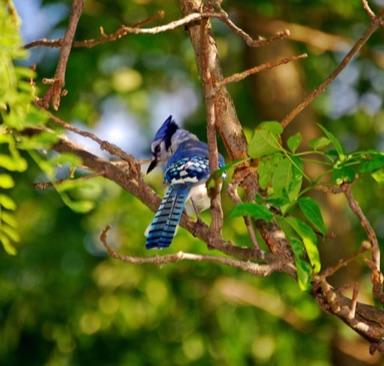} &
\includegraphics[width=0.158\linewidth]{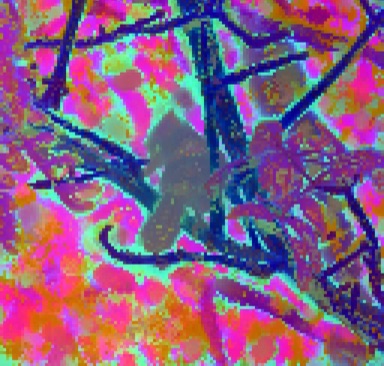} &
\includegraphics[width=0.158\linewidth]{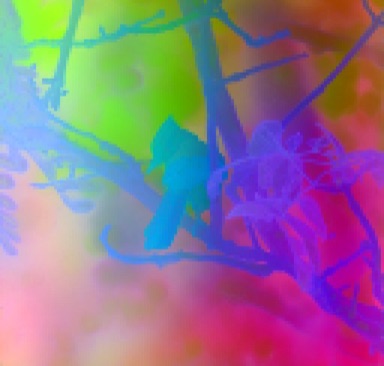} &
\includegraphics[width=0.158\linewidth]{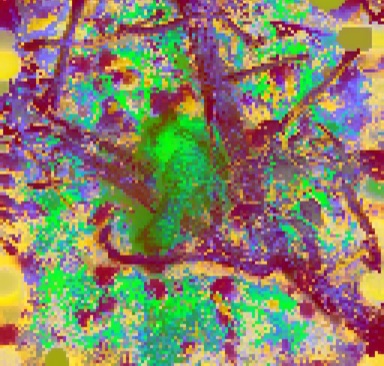} &
\includegraphics[width=0.158\linewidth]{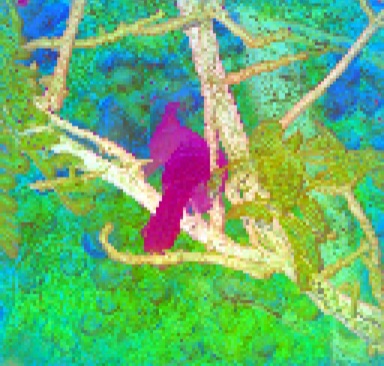} &
\includegraphics[width=0.158\linewidth]{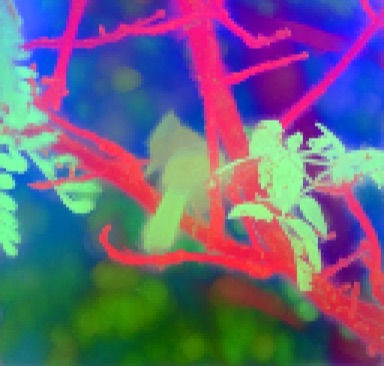} \\
\noalign{\vskip 1pt}
\includegraphics[width=0.158\linewidth]{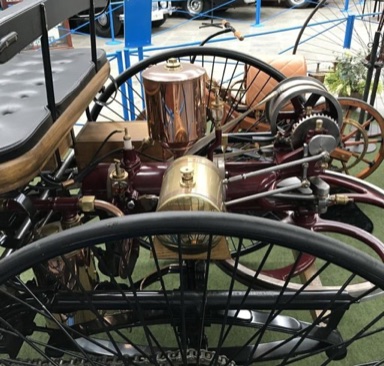} &
\includegraphics[width=0.158\linewidth]{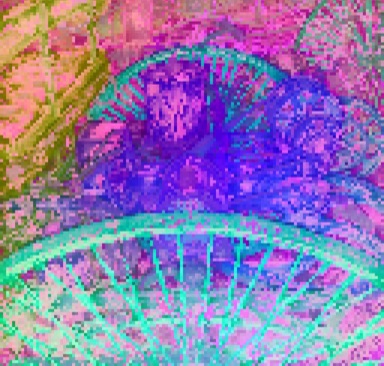} &
\includegraphics[width=0.158\linewidth]{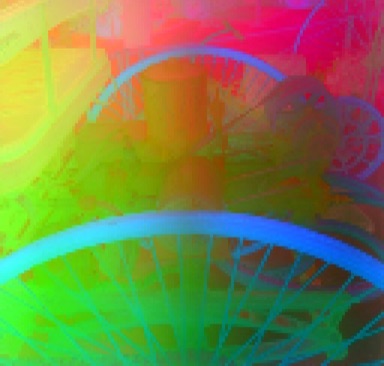} &
\includegraphics[width=0.158\linewidth]{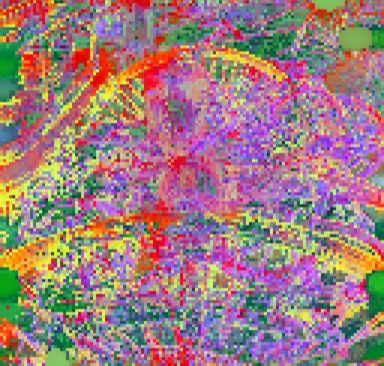} &
\includegraphics[width=0.158\linewidth]{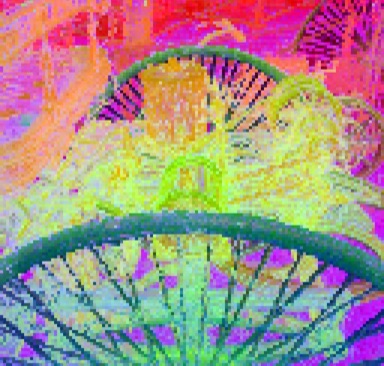} &
\includegraphics[width=0.158\linewidth]{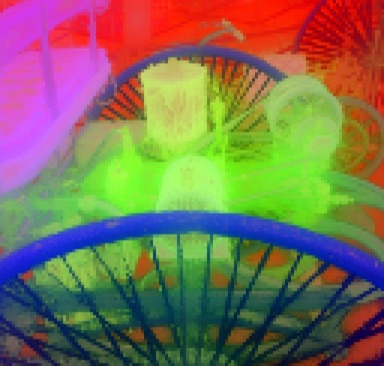} \\
\noalign{\vskip 1pt}
\includegraphics[width=0.158\linewidth]{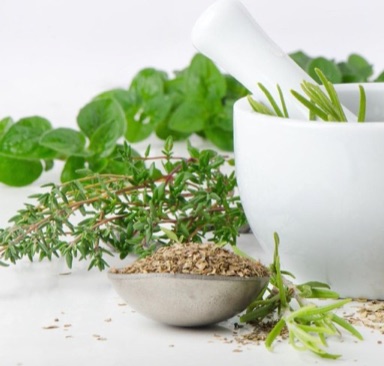} &
\includegraphics[width=0.158\linewidth]{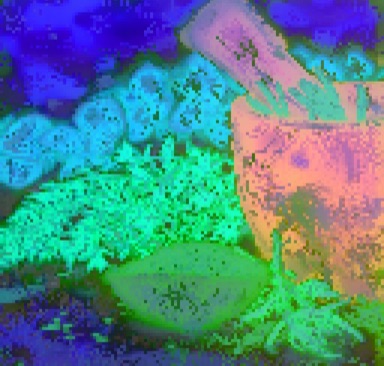} &
\includegraphics[width=0.158\linewidth]{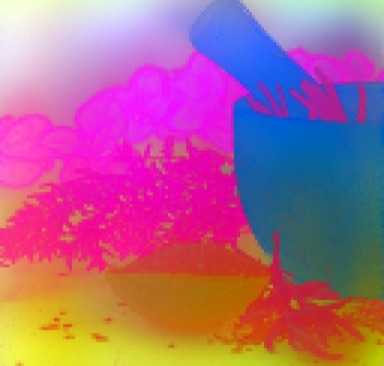} &
\includegraphics[width=0.158\linewidth]{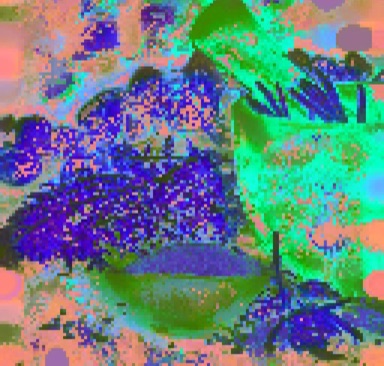} &
\includegraphics[width=0.158\linewidth]{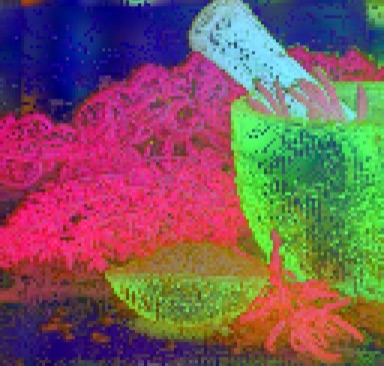} &
\includegraphics[width=0.158\linewidth]{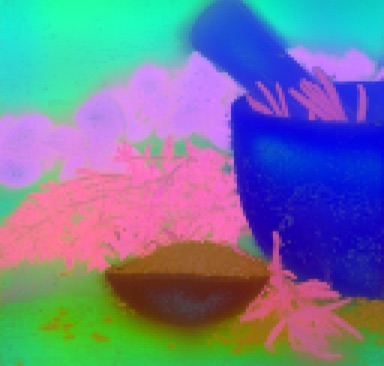} \\
\noalign{\vskip 1pt}
\includegraphics[width=0.158\linewidth]{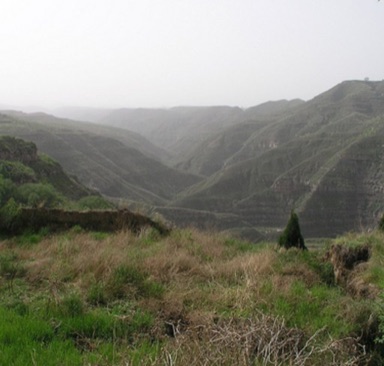} &
\includegraphics[width=0.158\linewidth]{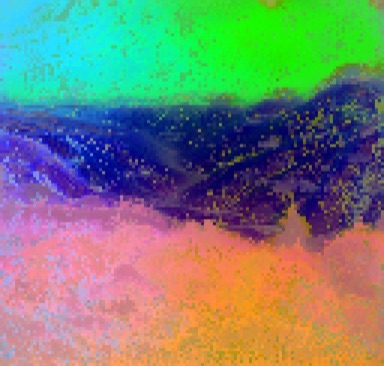} &
\includegraphics[width=0.158\linewidth]{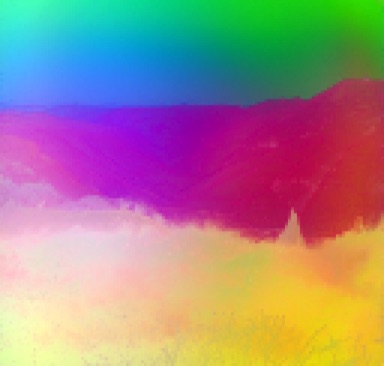} &
\includegraphics[width=0.158\linewidth]{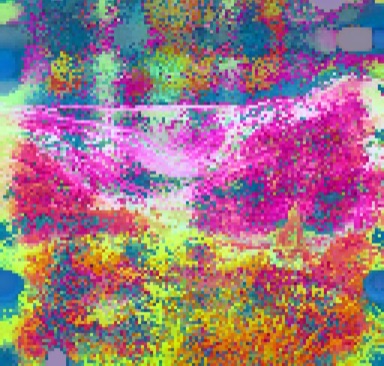} &
\includegraphics[width=0.158\linewidth]{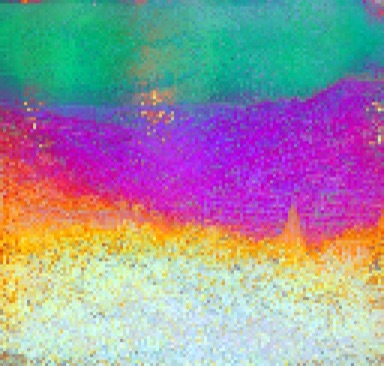} &
\includegraphics[width=0.158\linewidth]{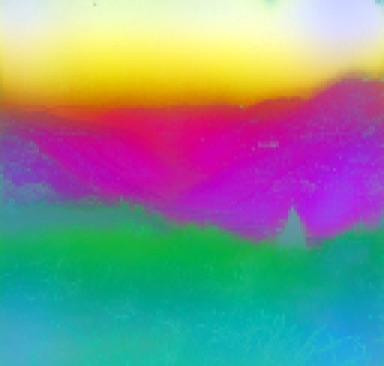} \\
\noalign{\vskip 1pt}
Images & DINOv2 & DINOv3 & SigLIP~2 & V-JEPA~2.1 & \method \\
\noalign{\vskip 1.5pt}
\end{tabular}
\caption{\textbf{PCA of frozen patch features.} \normalfont The top three
PCA components of the patch features are mapped to RGB, computed per image
for each model (one model per column). \method (rightmost) resolves objects
as coherent regions with crisp boundaries: individual cars and lane
structure in the traffic scene, hen silhouettes against the wire fence, the
winding contour of the snake, and fine structures such as flower stalks and
branches. In comparison, DINOv2 exhibits per-token speckle, SigLIP~2
degrades into blocky noise, and V-JEPA~2.1 lets background texture bleed
into the foreground regions.}
\label{fig:pca_comp}
\end{figure}
\section{Results}
\label{sec:exp}
In this section, we evaluate our pretrained vision foundation model,
\method, on various downstream tasks, following the standard protocol of
probing frozen features: linear probing on dense prediction and
classification tasks, and non-parametric probing ($k$-NN retrieval and label
propagation) on global and video tasks. The promising performance on public
benchmarks against the leading vision foundation models ({\em e.g.,}
DINOv2~\citep{oquab2023dinov2}, DINOv3~\citep{simeoni2025dinov3},
SigLIP~2~\citep{tschannen2025siglip} and V-JEPA~2.1~\citep{vjepa2p1}) lays a
solid foundation for spatial understanding, and culminates in LingBot-Depth 2.0 (Sec.~\ref{sec:lingbot_depth}).

\begin{table}
\caption{\textbf{Performance on dense visual tasks.} \normalfont
We report the Root Mean Squared Error (RMSE) for depth estimation (NYUv2, KITTI) and mean Intersection-over-Union (mIoU) for semantic segmentation (ADE20K, Cityscapes, VOC12). Following~\cite{simeoni2025dinov3}, ADE20K and VOC12 are evaluated at an input resolution of $448\times448$ for models with patch size 14 and $512\times512$ for those with patch size 16. Cityscapes, NYUv2, and KITTI use their default evaluation resolutions. With a 1B-parameter backbone, \method attains the best NYUv2 RMSE overall, ahead of the 7B-parameter DINOv3.}
\centering
\resizebox{.75\linewidth}{!}{%
\begin{tabular}{lccccccc}
\toprule
\multirow{2}{*}{\bf{Method}} & \multirow{2}{*}{\bf{Param.}} &
\multicolumn{2}{c}{\bf{Depth}} &
\multicolumn{3}{c}{\bf{Segmentation}} \\
\cmidrule(lr){3-4} \cmidrule(lr){5-7}
& &  NYUv2$\downarrow$ & KITTI$\downarrow$ & ADE20k & Citysc. & VOC \\
\midrule

\multicolumn{7}{l}{\textit{Models greater than 2B parameters}} \\
Web-DINO~\citep{fan2025webdino} & 7B/14 & 0.466 & 3.158 & 42.7 & 68.3 & 76.1 \\
DINOv3~\citep{simeoni2025dinov3} & 7B/16 & 0.309 & \bf{2.346} & \bf{55.9} & \bf{81.1} & \bf{86.6} \\
V-JEPA 2.1 ViT-G~\cite{vjepa2p1} & 2B/16 & \bf{0.307} & {2.461} & 47.9 & 73.5 & 85.0 \\
PEcore~\citep{bolya2025perception} & 2B/14 & 0.590 & 4.119 & 38.9 & 61.1 & 69.2 \\
PEspatial~\citep{bolya2025perception} & 2B/14 & 0.362 & 3.082 &49.3 & 73.2 & 82.7 \\
\midrule
\multicolumn{7}{l}{\textit{Models less than 2B parameters}} \\
AM-RADIOv2.5~\citep{ranzinger2024radio} & 1B/14 & 0.340 & 2.918 & 53.0 & 78.4 & 85.4 \\
InternVideo2-1B~\citep{wang2024internvideo2} & 1B/14  & 0.471 & 4.739 & 28.4 & 35.6 & 65.2 \\
SigLIP 2~\citep{tschannen2025siglip} & 1B/16 &  0.494 & 3.273 & 42.7 & 64.8 & 72.7 \\
DINOv2~\citep{oquab2023dinov2} & 1B/14  & 0.372 & 2.624 & 49.5 & 75.6 & 83.1 \\
DINOv3 ViT-H+~\citep{simeoni2025dinov3} & 0.8B/16 & 0.352 & 2.635 & \bf{54.8} & \underline{79.5} & \underline{85.8} \\
V-JEPA2 ViT-g~\citep{assran2025v} & 1B/16 & 0.642 & 4.650 & 24.4 & 45.9 & 63.9 \\
V-JEPA 2.1 ViT-g~\cite{vjepa2p1}  & 1B/16 & \underline{0.350} & \underline{2.601} & 47.8 & 71.8 & 84.7 \\
\midrule
LingBot-Vision ViT-g & 1B/16 & {\bf 0.296} & {\bf 2.552} & \underline{53.5} & {\bf 79.6} & {\bf 87.5}
\\
\bottomrule
\end{tabular}
}

\label{tab:segmentation_depth_extended}
\end{table}
\subsection{Robust Dense Representation of Images}
\label{sec:exp:dense}

\subsubsection{PCA Components of Patch Features}
Before any probing, the structure of the
frozen features is already visible to the naked eye.
Fig.~\ref{fig:pca_comp} maps the top three PCA components of the patch
features to RGB for \method and the leading vision foundation models,
computed per image. Across scenes, \method resolves objects as coherent,
sharply bounded regions: individual cars and the lane layout of the highway,
hen silhouettes cleanly separated from the wire fence, the winding contour
of the snake, and fine structures such as flower stalks around the marmot.
The comparison also exposes characteristic failure modes of the baselines:
DINOv2 features carry per-token speckle, SigLIP~2 collapses into blocky
noise away from salient objects, and V-JEPA~2.1 lets background texture
bleed into foreground regions. These qualitative differences anticipate the
quantitative gaps below: boundary-faithful features are precisely what
pixel-dense readouts consume.

\subsubsection{Linear Probing on Depth Estimation}
\paragraph{Evaluation protocol.} We follow the standard linear probing
protocol: the ViT backbone is frozen, and only the decoder is trained on
the training split of each target dataset. Instead of practices that add
decoding capacity ({\em e.g.,} a DPT head~\citep{ranftl2021dpt} or
multi-layer feature aggregation), we use a single linear layer on the
frozen patch tokens as the decoder for all pretrained vision transformers,
so the comparison attributes performance to the features rather than to the
readout. NYUv2~\citep{nyuv2} and KITTI~\citep{kitti} are evaluated at their
default resolutions and reported in RMSE.

\paragraph{Results.} Table~\ref{tab:segmentation_depth_extended} groups the
competitors by parameter count. With a 1B-parameter ViT-g/16 backbone,
\method attains the best NYUv2 RMSE of the entire table (0.296), ahead of
the 7B-parameter DINOv3 (0.309) and the 2B-parameter V-JEPA~2.1 (0.307),
with 7$\times$ and 2$\times$ fewer parameters respectively. The margin over
models at comparable scale is substantial: 13\% lower RMSE than the best
same-scale competitor AM-RADIOv2.5 (0.340), 16\% lower than DINOv3 ViT-H+
(0.352), which is distilled from the 7B teacher, and 20\% lower than DINOv2
(0.372). The strongest video-pretrained baseline at equal size,
V-JEPA~2.1 ViT-g (0.350), remains 15\% behind. On KITTI, \method obtains
the best RMSE among models below 2B parameters (2.552), surpassed only by
the 7B DINOv3 (2.346) and the 2B V-JEPA~2.1 (2.461). Notably, \method
achieves these results with patch size 16, a coarser token grid than the
patch-14 competitors at equal input resolution.

\paragraph{Analysis.} A linear decoder cannot compensate for deficiencies
of the representation; it can only read out what the frozen features
already encode. Accurate depth under this protocol therefore
requires features that are smooth within a surface and change sharply
across occlusion boundaries, exactly where depth discontinuities
concentrate. This is the regime that the boundary-oriented pretraining of
\method targets, and it is visible in Fig.~\ref{fig:pca_comp}: object
interiors map to coherent regions while transitions stay crisp. Depth is
consequently the task where \method holds the largest margin, and the
margin is wider on NYUv2 than on KITTI, where the spread across all
backbones compresses: outdoor driving scenes are dominated by road and sky
regions with few nearby occlusion boundaries, so boundary fidelity
contributes less of the error budget.

\subsubsection{Linear Probing on Semantic Segmentation}
\paragraph{Evaluation protocol.} The same single-linear-layer decoder is
trained per dataset on the frozen patch tokens. Following the DINOv3
protocol, ADE20K~\citep{ade20k} and VOC12~\citep{voc} are evaluated at
$448\times448$ for models with patch size 14 and $512\times512$ for patch
size 16, so that all models produce the same patch-token resolution;
Cityscapes~\citep{cityscapes} uses its default evaluation resolution. We report the mean Intersection-over-Union (mIoU) against the ground-truth annotations.

\paragraph{Results.} On ADE20K, Cityscapes and VOC12 (Table~\ref{tab:segmentation_depth_extended}), \method is on par with the distilled DINOv3 ViT-H+: it trails by 1.3 mIoU on ADE20K (53.5 vs. 54.8) while matching it on Cityscapes and leading on VOC12. Against the standard self-supervised reference at equal size, DINOv2, the improvement is a consistent 4 mIoU or more on all three benchmarks ({\em e.g.,} 53.5 vs. 49.5 on ADE20K), and the equal-size video-pretrained V-JEPA~2.1 ViT-g trails by 5.7 mIoU on ADE20K. \method also surpasses AM-RADIOv2.5 on all three benchmarks, although the latter agglomerates several teachers including segmentation-specialized ones, and it exceeds the spatially post-trained PEspatial by 4.2 mIoU on ADE20K. The remaining gap is only to the DINOv3 family (2.4 mIoU on ADE20K to the 7B model), whose members benefit from distillation and dedicated dense-feature objectives.  We emphasize that \method is pretrained from scratch with a single self-supervised objective, whereas the strongest dense baselines rely on distillation from a 7B teacher (DINOv3 ViT-H+) or task-oriented post-training (PEspatial).

\paragraph{Analysis.} Segmentation rewards two properties jointly: patches of the same semantic region must map to nearby features, and the feature transition must land exactly on the class boundary, since mIoU errors concentrate along contours and thin structures. The PCA maps of Fig.~\ref{fig:pca_comp} show that \method provides both, with coherent object interiors and crisp transitions. The comparison also separates objective families: language-aligned pretraining (SigLIP~2, 42.7 mIoU on ADE20K) optimizes image-level alignment at the expense of spatial detail, and video-predictive pretraining recovers dense structure only after its refinements (V-JEPA~2 to V-JEPA~2.1, 24.4 to 47.8 mIoU). Boundary-oriented image pretraining reaches distillation-level segmentation quality without a larger teacher.

\subsection{Video Understanding}
\label{sec:exp:vos}
We assess the temporal consistency of the frozen features in two ways:
quantitatively, with training-free video object segmentation on standard
benchmarks, and qualitatively, by tracking individual boundary tokens
through self-captured videos and visualizing the similarity responses of
the queried tokens over time.

\subsubsection{Video Object Segmentation}
\paragraph{Evaluation protocol.} We evaluate training-free video object
segmentation on DAVIS-2017~\citep{davis} and YouTube-VOS~\citep{ytvos}: the ground-truth mask of the
first frame is propagated to subsequent frames by top-$k$ attention over
the frozen patch features, without any fine-tuning. Videos are resized so
that the shorter side is 480 px for models with patch size 16 and 420 px
for those with patch size 14, and we report the
$\mathcal{J}\&\mathcal{F}$-Mean.

\paragraph{Results.} As reported in Table~\ref{tab:vos}, \method reaches
70.0 and 73.5 $\mathcal{J}\&\mathcal{F}$ on DAVIS and YouTube-VOS
respectively, on par with DINOv3 ViT-H+ (71.1 and 74.0) and the 7B DINOv3
(71.1 and 74.1), and the best among all remaining models at any scale. At
equal size, the margin over DINOv2 is 6.1 and 7.9 points, and the
video-pretrained V-JEPA~2.1 ViT-g trails by 1.9 and 1.2 points despite
being trained on video.

\begin{table}[t]
\caption{\textbf{Video object segmentation with frozen features.}
\normalfont We report the $\mathcal{J}\&\mathcal{F}$-Mean on DAVIS-2017 and
YouTube-VOS under training-free label propagation, with the shorter image
side set to 480 px for models with patch size 16 and 420 px for those with
patch size 14.}
\centering
\begin{tabular}{lccc}
\toprule
\bf{Method} & \bf{Param.} & DAVIS-S & YT VOS-S \\
\midrule
\multicolumn{4}{l}{\textit{Models greater than 2B parameters}} \\
Web-DINO~\citep{fan2025webdino} & 7B/14 & 57.2 & 43.9 \\
DINOv3~\citep{simeoni2025dinov3} & 7B/16 & \bf{71.1} & \bf{74.1} \\
V-JEPA 2.1 ViT-G~\cite{vjepa2p1} & 2B/16 & 69.0 & 72.7 \\
PEcore~\citep{bolya2025perception} & 2B/14 & 48.2 & 34.7 \\
PEspatial~\citep{bolya2025perception} & 2B/14 & 68.4 & 68.5 \\
\midrule
\multicolumn{4}{l}{\textit{Models less than 2B parameters}} \\
AM-RADIOv2.5~\citep{ranzinger2024radio} & 1B/14 & 66.5 & 70.1 \\
InternVideo2-1B~\citep{wang2024internvideo2} & 1B/14 & 50.6 & 51.2 \\
SigLIP 2~\citep{tschannen2025siglip} & 1B/16 & 56.1 & 52.0 \\
DINOv2~\citep{oquab2023dinov2} & 1B/14 & 63.9 & 65.6 \\
DINOv3 ViT-H+~\citep{simeoni2025dinov3} & 0.8B/16 & \bf{71.1} & \bf{74.0} \\
V-JEPA2 ViT-g~\citep{assran2025v} & 1B/16 & 52.5 & 53.7 \\
V-JEPA 2.1 ViT-g~\cite{vjepa2p1} & 1B/16 & 68.1 & 72.3 \\
\midrule
LingBot-Vision ViT-g (Ours) & 1B/16 & 70.0 & 73.5 \\
\bottomrule
\end{tabular}
\label{tab:vos}
\end{table}

\subsubsection{Boundary Token Tracking}
\begin{figure}[!t]
\centering
\setlength{\tabcolsep}{0.5pt}
\renewcommand{\arraystretch}{0}
\begin{tabular}{cc}
\rotatebox{90}{\scriptsize\makebox[0.14\linewidth][c]{$t\,{=}\,0$\,s}} &
\includegraphics[width=\linewidth]{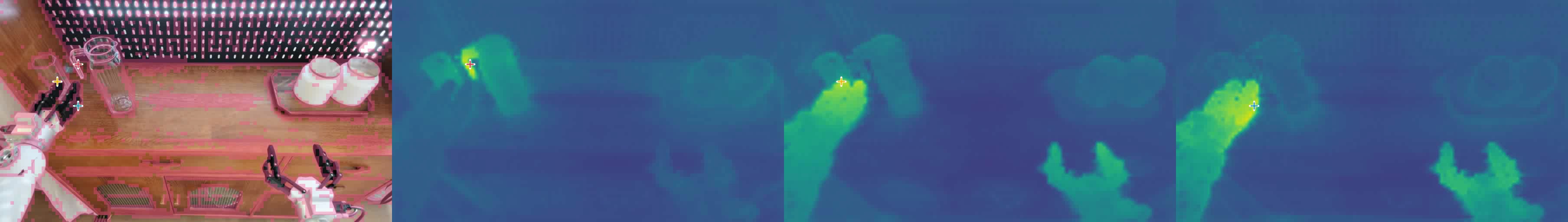} \\
\noalign{\vskip 1pt}
\rotatebox{90}{\scriptsize\makebox[0.14\linewidth][c]{$t\,{\approx}\,100$\,s}} &
\includegraphics[width=\linewidth]{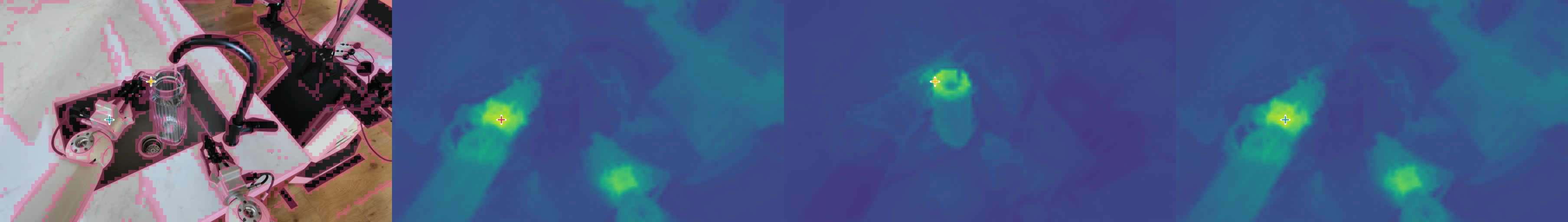} \\
\noalign{\vskip 1pt}
\noalign{\vskip 3pt}
\rotatebox{90}{\scriptsize\makebox[0.14\linewidth][c]{$t\,{=}\,0$\,s}} &
\includegraphics[width=\linewidth]{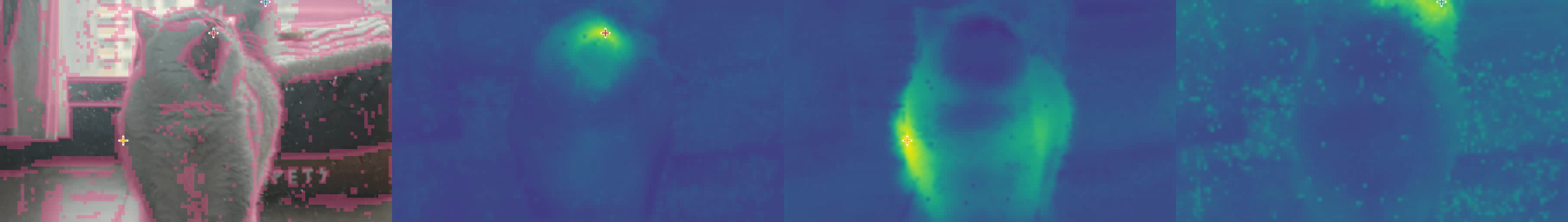} \\
\noalign{\vskip 1pt}
\rotatebox{90}{\scriptsize\makebox[0.14\linewidth][c]{$t\,{\approx}\,3.5$\,s}} &
\includegraphics[width=\linewidth]{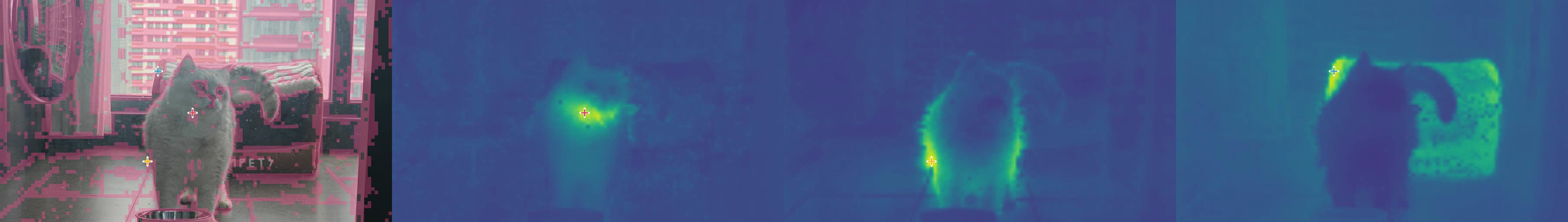} \\
\noalign{\vskip 1pt}
\noalign{\vskip 3pt}
\rotatebox{90}{\scriptsize\makebox[0.14\linewidth][c]{$t\,{\approx}\,3$\,s}} &
\includegraphics[width=\linewidth]{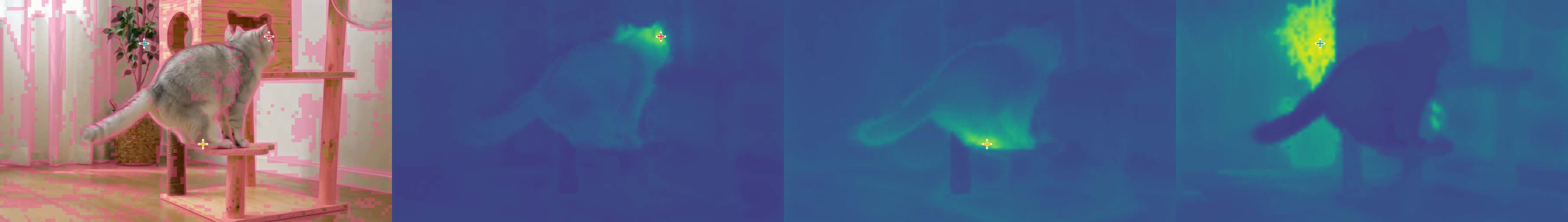} \\
\noalign{\vskip 1pt}
\rotatebox{90}{\scriptsize\makebox[0.14\linewidth][c]{$t\,{=}\,10$\,s}} &
\includegraphics[width=\linewidth]{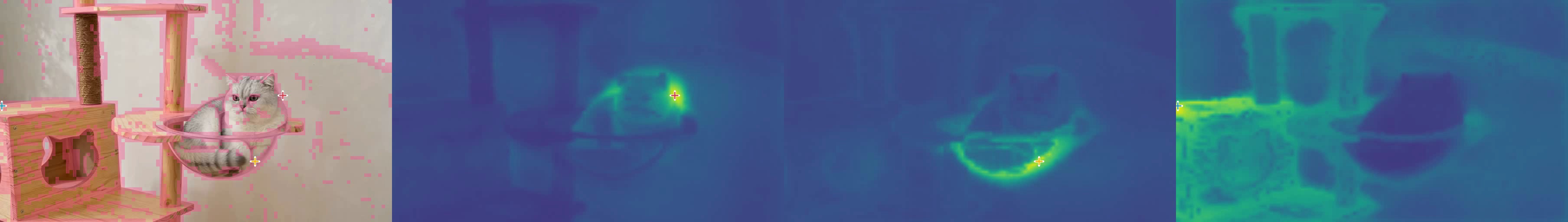} \\
\noalign{\vskip 1pt}
\end{tabular}
\caption{\textbf{Boundary-token tracking on three videos.}
\normalfont Three boundary-token queries are seeded in the first frame of each video (crosses in the left panels) and tracked by cosine similarity of the frozen patch features over each frame's boundary tokens; the three heatmap panels show the similarity responses of the three queries. Top pair: a manipulation episode seen from a robot's head camera, where the queries hold the gripper fingers and the pitcher throughout the episode.  Middle pair: a self-captured cat video, where the queries on the head and flank follow a full turn of the body while the query on the window frame stays put. Bottom pair: a camera pan through a home, where the queries track the cat and the cat-tree platform through a large viewpoint change and re-acquire the plant as it re-enters the view. No fine-tuning or temporal supervision is used.}
\label{fig:token_track}
\end{figure}

Beyond aggregate benchmarks, we probe temporal consistency at the level of individual tokens. Fig.~\ref{fig:token_track} tracks three boundary-token queries through three videos of different nature: a manipulation episode seen from a robot's head camera, a self-captured cat video, and a camera pan through a home. At every frame, each query is matched by cosine similarity of the frozen patch features over the boundary tokens of that frame, within a local window around its previous position, and the query feature is updated from its best match while staying anchored to its first-frame template, so the tracker follows appearance changes without drifting. The queries stay locked in all three regimes: on the robot, the gripper fingers and the manipulated pitcher are held through the whole episode; on the cat, the head and flank queries follow a full turn of the body while the query on the static window frame stays put; and under the camera pan, the queries survive a large viewpoint change and re-acquire the plant when it re-enters the view. Throughout, the similarity responses remain selective rather than diffusing over the scene. Boundary tokens thus behave as stable, trackable entities in the feature space of \method, obtained without any temporal supervision.

\subsection{Strong Distilled Models on both Global and Dense Tasks}
\label{sec:exp:global}
\label{sec:exp:distill}

We complete the picture with image-level recognition, evaluating both the giant teacher and the distilled \method family, since classification is also the axis along which we distill.

\subsubsection{Linear and $k$-NN Probing on ImageNet}

\begin{table}
\centering
\small
\vspace{-3mm}
\caption{\textbf{ImageNet-1k accuracy of the flagship models.}
\normalfont Linear-probe and $k$-NN top-1 accuracy at giant scale; the
distilled families are compared in
Table~\ref{tab:lbs_distill_global_dense}.}
\label{tab:giant_global}
\begin{tabular}{lcc}
\toprule
Model & IN-1K Linear & IN-1K $k$-NN\\
\midrule
LingBot-Vision (Ours) & 86.32 & 83.39 \\
DINOv2~\cite{oquab2023dinov2}         & 87.00 & 83.68 \\
DINOv3~\cite{simeoni2025dinov3}         & 87.87 & 85.68 \\
Franca~\cite{venkataramanan2025franca}         & 84.50 & 81.11 \\
SigLIP2~\cite{tschannen2025siglip}        & 87.33 & 84.75 \\
\bottomrule
\end{tabular}
\end{table}
\paragraph{Evaluation protocol.} We evaluate ImageNet-1K~\citep{imagenet} classification on frozen features with two readouts: a linear classifier trained on the training split, and a training-free $k$-NN classifier. Evaluation is token-aligned: input resolutions are set per patch size so that all models produce the same number of patch tokens, consistent with the dense protocols above. The linear probe follows the standard grid-search practice: SGD with a momentum of 0.9, trained for 10 epochs with a batch size of 1024, with the learning rate swept over $\{1, 2, 5\} \times 10^{k}$ for $k \in \{-4, \dots, 0\}$ and the weight decay over $\{0, 10^{-5}\}$, selecting the best combination on the ImageNet-1K validation set; training uses Inception-style random resized crops, and evaluation uses $224\times224$ images.

\paragraph{Results.} At giant scale (Table~\ref{tab:giant_global}), \method obtains 86.32 (linear) and 83.39 ($k$-NN) top-1 accuracy, on par with DINOv2's flagship model (87.00 and 83.68) and behind DINOv3-7B (87.87 and 85.68) and SigLIP~2 (87.33 and 84.75). The remaining gap to DINOv3-7B thus concentrates on image-level recognition, while the dense comparisons above favor \method: a trade-off consistent with our boundary-oriented pretraining, which invests model capacity in localized structure rather than image-level invariances. 

\begin{table}[t]
\centering
\caption{
Comparison of LingBot-Vision distilled models with representative pretrained vision backbones.
Global tasks report ImageNet-1k linear probing and KNN top-1 accuracy, while dense tasks report depth RMSE$\downarrow$, and semantic segmentation mIoU. We match the patch token resolution of models with different patch size.
}
\label{tab:lbs_distill_global_dense}
\resizebox{.8\textwidth}{!}{
\begin{tabular}{llccccccc}
\toprule
\multirow{2}{*}{Size} & \multirow{2}{*}{Model}
& \multicolumn{2}{c}{Global Tasks}
& \multicolumn{5}{c}{Dense Tasks} \\
\cmidrule(lr){3-4}\cmidrule(lr){5-9}
& & IN1k-Linear & IN1k-KNN & NYU$\downarrow$ & KITTI$\downarrow$ & ADE & VOC & City \\
\midrule
\multirow{6}{*}{L}
& DINOv2~\cite{oquab2023dinov2}         & \underline{86.43} & \underline{83.82} & 0.411 & 3.243 & 48.63 & 83.54 & 74.51 \\
& DINOv3~\cite{simeoni2025dinov3}         & \textbf{87.31} & \textbf{85.27} & 0.351 & 2.643 & \textbf{55.00} & \textbf{88.11} & \textbf{79.28} \\
& SigLIP2~\cite{tschannen2025siglip}        & 85.67 & 82.91 & 0.516 & 3.641 & 42.99 & 75.55 & 62.51 \\
& Franca~\cite{venkataramanan2025franca}         & 82.50 & 71.70 & 0.425 & 3.613 & 45.93 & 84.42 & 72.42 \\
& V-JEPA2.1~\cite{vjepa2p1}        & -- & -- & \underline{0.346} & \underline{2.617} & 46.29 & 84.81 & 71.58 \\
& LingBot-Vision & 86.38 & 83.62 & \textbf{0.310} & \textbf{2.574} & \underline{52.75} & \underline{87.40} & \underline{78.75} \\
\midrule
\multirow{6}{*}{B}
& DINOv2~\cite{oquab2023dinov2}         & 84.27 & 82.17 & 0.429 & 3.416 & 48.25 & 83.81 & 74.19 \\
& DINOv3~\cite{simeoni2025dinov3}         & \underline{84.79} & \textbf{83.21} & \underline{0.371} & \underline{2.826} & \textbf{51.74} & \textbf{88.20} & \textbf{77.30} \\
& SigLIP2~\cite{tschannen2025siglip}        & 82.47 & 79.47 & 0.539 & 3.847 & 40.49 & 73.50 & 60.12 \\
& Franca~\cite{venkataramanan2025franca}         & 78.45 & 63.45 & 0.462 & 3.790 & 42.69 & 82.79 & 71.65 \\
& V-JEPA2.1~\cite{vjepa2p1}       & -- & -- & 0.410 & 2.883 & 41.38 & 78.00 & 64.99 \\
& LingBot-Vision (Ours) & \textbf{85.05} & \underline{82.32} & \textbf{0.339} & \textbf{2.793} & \underline{51.44} & \underline{87.10} & \underline{77.17} \\
\midrule
\multirow{3}{*}{S}
& DINOv2~\cite{oquab2023dinov2}         & \underline{80.76} & \underline{79.11} & 0.447 & \underline{3.568} & 44.86 & 82.34 & \underline{71.44} \\
& DINOv3~\cite{simeoni2025dinov3}         & 80.37 & \textbf{79.33} & \underline{0.405} & \textbf{2.851} & \underline{46.51} & \textbf{84.30} & \textbf{73.38} \\
& LingBot-Vision (Ours) & \textbf{82.22} & 78.98 & \textbf{0.383} & 3.784 & \textbf{47.01} & \underline{84.17} & 64.34 \\
\bottomrule
\end{tabular}
}
\end{table}
\subsubsection{Distilled Models Across Scales}
\paragraph{Setup.} To serve different deployment budgets, we distill the
giant \method teacher into ViT-L, ViT-B and ViT-S students, and probe them
with the same global and dense protocols
(Table~\ref{tab:lbs_distill_global_dense}).

\paragraph{Results.} The distilled family preserves the profile of the
teacher at every size. At ViT-L, \method matches DINOv2 on global accuracy
(86.38 vs. 86.43 linear) while reducing the NYUv2 RMSE by 25\% (0.310
vs. 0.411); against DINOv3-L it concedes 0.9 points of linear accuracy but
leads depth by 12\% on NYUv2 and stays within 0.5 to 2.3 mIoU across the
segmentation benchmarks. Notably, the 0.3B student attains a NYUv2 RMSE on
par with the 7B DINOv3 of
Table~\ref{tab:segmentation_depth_extended} (0.310 vs. 0.309) under the
same protocol, with roughly 23$\times$ fewer parameters. At ViT-B,
\method obtains the best linear accuracy in its size class (85.05)
together with the best depth on both datasets (0.339 NYUv2, 2.793 KITTI),
trailing DINOv3-B by at most 0.3 mIoU on ADE20K and Cityscapes. At ViT-S,
it again leads linear probing (82.22), NYUv2 (0.383) and ADE20K (47.01);
the smallest student stays within 0.4 points on $k$-NN and VOC but falls
behind on KITTI and Cityscapes. Across sizes, the video-pretrained V-JEPA2.1 trails on every
dense benchmark, SigLIP~2 remains close on classification but far behind
on dense tasks, and Franca trails on both axes.

\paragraph{Analysis.} Two observations stand out. First, the dense
advantage transfers through distillation essentially intact: the students
inherit the boundary-faithful patch features that make the teacher strong,
while the residual gaps to DINOv3 concentrate on image-level accuracy,
mirroring the giant-scale comparison above. Second, the family forms a
practical accuracy-compute frontier: at every deployment budget, the
\method student offers the best available depth accuracy in its class
with classification within a point of the best competitor, so downstream
users do not trade dense quality for model size.

The frozen-feature results above, together with the encoder-initialization
study of Table~\ref{tab:dc_encoder_init}, establish \method as a strong
spatial backbone. Sec.~\ref{sec:lingbot_depth} builds on it to develop
LingBot-Depth 2.0 for metric depth completion.

\section{Meet LingBot-Depth 2.0}
\label{sec:lingbot_depth}
Based on the success of \method on general vision pretraining, it is time to upgrade our previous LingBot-Depth~\cite{lingbot-depth} for depth completion, a robotics-oriented perception task, with \method as its spatial-perception-native backbone. To be self-contained, we first briefly recap how LingBot-Depth was trained in its previous version, and then describe our tweaks toward stronger performance. At a glance, LingBot-Depth 2.0 comes in two sizes, built on the ViT-L/16 and ViT-g/16 backbones from our \method pretraining, and we explore data scaling by growing the training set from the publicly released 3M samples to a newly curated 150M collection, setting the leading performance on 14 benchmarks spanning different depth patterns and camera types.

\begin{table}[t]
    \centering
    \caption{\textbf{Encoder initialization study for masked depth
    modeling.} \normalfont The exact same MDM pipeline is trained from
    different encoder initializations using the {\bf same data} and evaluated on block-masked and
    sparse input depth (top) and on raw captures of real depth cameras (bottom). Each cell reports RMSE$\downarrow$\,/\,$D_{105}\uparrow$;
    the best result per column within each scale group is in bold. The
    \method initialization is the strongest starting point at ViT-L
    across the board and on the majority of benchmarks at ViT-g, where
    DINOv2 keeps an edge on the Hammer captures.}
    \label{tab:dc_encoder_init}
    \resizebox{\columnwidth}{!}{
    \begin{tabular}{ll|cccccccc}
    \toprule
    & & \multicolumn{4}{c}{\bf Block Mask} & \multicolumn{4}{c}{\bf Sparse} \\
    \cmidrule(lr){3-6}\cmidrule(lr){7-10}
    \bf Init & \bf Arch & DIODE-In & DIODE-Out & Ibims-1 & NYU & VOID & Ibims-1 & NYU & ETH3D \\
    \midrule
    DINOv2~\cite{oquab2023dinov2} & ViT-L/14 & 0.152/0.981 & 3.192/0.775 & 0.192/0.942 & 0.169/0.920 & 0.214/0.732 & 0.167/0.944 & 0.139/0.930 & {\bf 0.385}/0.865 \\
    DINOv3~\cite{simeoni2025dinov3} & ViT-L/16 & 0.114/0.988 & 3.065/0.735 & 0.189/0.945 & 0.154/0.922 & {\bf 0.190}/0.885 & 0.167/0.949 & 0.139/0.929 & 0.460/0.842 \\
    \method & ViT-L/16 & {\bf 0.094}/{\bf 0.990} & {\bf 2.771}/{\bf 0.794} & {\bf 0.167}/{\bf 0.953} & {\bf 0.145}/{\bf 0.930} & 0.199/{\bf 0.887} & {\bf 0.167}/{\bf 0.952} & {\bf 0.136}/{\bf 0.932} & 0.414/{\bf 0.868} \\
    \midrule
    DINOv2 & ViT-g/14 & 0.118/0.990 & 2.778/{\bf 0.809} & 0.180/0.952 & 0.178/0.922 & 0.203/0.810 & 0.156/0.957 & 0.142/0.932 & {\bf 0.361}/{\bf 0.911} \\
    \method & ViT-g/16 & {\bf 0.083}/{\bf 0.993} & {\bf 2.734}/0.801 & {\bf 0.166}/{\bf 0.959} & {\bf 0.148}/{\bf 0.935} & {\bf 0.182}/{\bf 0.916} & {\bf 0.142}/{\bf 0.964} & {\bf 0.125}/{\bf 0.942} & 0.365/0.911 \\
    \midrule
    & & \multicolumn{3}{c}{\bf Hammer} & \multicolumn{2}{c}{\bf ClearGrasp} & \multicolumn{3}{c}{\bf LingBot} \\
    \cmidrule(lr){3-5}\cmidrule(lr){6-7}\cmidrule(lr){8-10}
    \bf Init & \bf Arch & D435 & L515 & ToF & D415 & D435 & D415 & D435 & D455 \\
    \midrule
    DINOv2~\cite{oquab2023dinov2} & ViT-L/14 & 0.034/0.860 & 0.021/{\bf 0.960} & 0.030/0.900 & 0.012/0.958 & 0.015/0.915 & 0.295/0.932 & 0.400/{\bf 0.956} & 0.454/0.900 \\
    DINOv3~\cite{simeoni2025dinov3} & ViT-L/16 & 0.033/0.871 & 0.022/0.946 & 0.029/0.878 & 0.012/0.968 & 0.012/0.953 & 0.289/0.930 & 0.377/0.955 & {\bf 0.414}/0.903 \\
    \method & ViT-L/16 & {\bf 0.031}/{\bf 0.892} & {\bf 0.020}/0.951 & {\bf 0.027}/{\bf 0.928} & {\bf 0.010}/{\bf 0.981} & {\bf 0.010}/{\bf 0.985} & {\bf 0.271}/{\bf 0.936} & {\bf 0.373}/0.955 & 0.415/{\bf 0.906} \\
    \midrule
DINOv2~\cite{oquab2023dinov2} & ViT-g/14 & {\bf 0.031}/{\bf 0.910} & {\bf 0.021}/{\bf 0.966} & {\bf 0.024}/{\bf 0.954} & \underline{0.013}/\underline{0.946} & \underline{0.018}/\underline{0.861} & \underline{0.260}/\underline{0.944} & \underline{0.373}/\underline{0.962} & \underline{0.447}/\underline{0.918} \\
    \method & ViT-g/16 & \underline{0.031}/\underline{0.905} & \underline{0.022}/\underline{0.950} & \underline{0.027}/\underline{0.951} & {\bf 0.011}/{\bf 0.973} & {\bf 0.015}/{\bf 0.928} & {\bf 0.254}/{\bf 0.951} & {\bf 0.359}/{\bf 0.965} & {\bf 0.409}/{\bf 0.930} \\
    \bottomrule
    \end{tabular}
    }
\end{table}

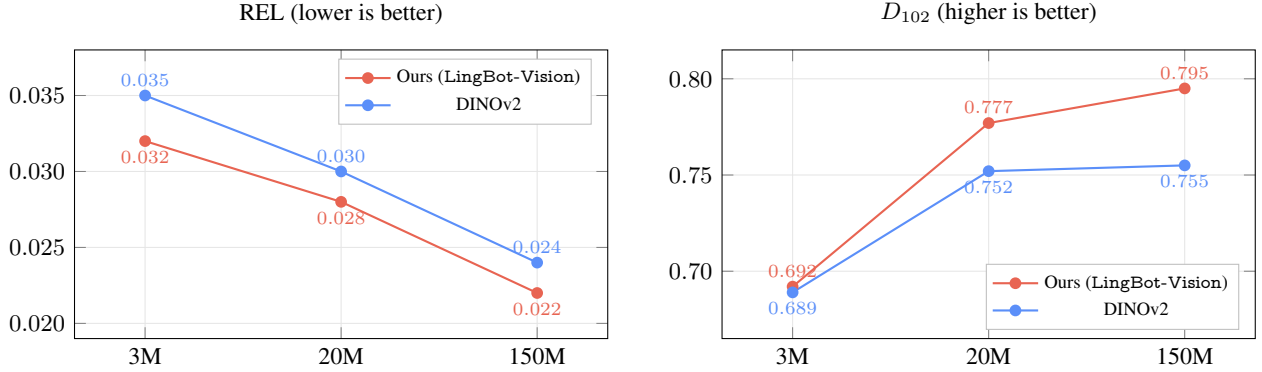
\begin{figure}[!t]
    \centering
    \pgfplotsset{dsaxis/.style={
        width=0.52\linewidth, height=5.4cm,
        symbolic x coords={3M,20M,150M}, xtick=data,
        enlarge x limits=0.18,
        grid=major, grid style={black!10},
        tick label style={font=\small},
        title style={font=\small},
        legend style={font=\scriptsize, draw=black!25, fill=white,
                      inner sep=2pt},
        every axis plot/.append style={thick, mark=*, mark size=1.8pt},
        nodes near coords,
        every node near coord/.append style={font=\scriptsize,
            /pgf/number format/fixed, /pgf/number format/fixed zerofill,
            /pgf/number format/precision=3},
        clip=false,
    }}
    \begin{tikzpicture}
    \begin{axis}[dsaxis, title={REL (lower is better)},
        legend pos=north east, ymin=0.019, ymax=0.038,
        scaled y ticks=false,
        y tick label style={/pgf/number format/fixed,
            /pgf/number format/fixed zerofill,
            /pgf/number format/precision=3}]
    \addplot[pie4, mark options={fill=pie4},
        nodes near coords style={anchor=north}] coordinates
        {(3M,0.032) (20M,0.028) (150M,0.022)};
    \addplot[pie1, mark options={fill=pie1},
        nodes near coords style={anchor=south}] coordinates
        {(3M,0.035) (20M,0.030) (150M,0.024)};
    \legend{Ours (\method), DINOv2}
    \end{axis}
    \end{tikzpicture}\hfill
    \begin{tikzpicture}
    \begin{axis}[dsaxis, title={$D_{102}$ (higher is better)},
        legend pos=south east, ymin=0.665, ymax=0.815,
        y tick label style={/pgf/number format/fixed,
            /pgf/number format/fixed zerofill,
            /pgf/number format/precision=2}]
    \addplot[pie4, mark options={fill=pie4},
        nodes near coords style={anchor=south}] coordinates
        {(3M,0.692) (20M,0.777) (150M,0.795)};
    \addplot[pie1, mark options={fill=pie1},
        nodes near coords style={anchor=north}] coordinates
        {(3M,0.689) (20M,0.752) (150M,0.755)};
    \legend{Ours (\method), DINOv2}
    \end{axis}
    \end{tikzpicture}
    \caption{\textbf{Scaling the curated training data.} \normalfont
    Depth completion quality as the MDM training set grows from 3M to 20M and 150M samples, for the \method and DINOv2 encoder initializations under the same pipeline. Left: mean absolute relative error (REL, lower is better); right: strict threshold accuracy $D_{102}$ (higher is better). Both initializations improve monotonically with data, but they are not interchangeable: the two start nearly tied on $D_{102}$ at 3M, and the gap widens with scale as the DINOv2 curve saturates beyond 20M (0.752 to 0.755) while the \method curve keeps improving (0.777 to 0.795). More data amplifies, rather than washes out, the advantage of the spatial-perception-native initialization.}
    \label{fig:data_size}
\end{figure}

\begin{table}[]
    \centering
    \caption{\textbf{Depth completion on real sensor captures.}
    \normalfont Raw, incomplete depth maps from three camera families
    (HAMMER: D435, L515, ToF; ClearGrasp: D415, D435; LingBot: D415,
    D435, D455) are completed by each method. Each cell reports
    RMSE$\downarrow$\,/\,$D_{105}\uparrow$; the best result per column
    is in bold and the second best is underlined. LingBot-Depth 2.0
    leads on six of the eight configurations and is strongest on the
    transparent-object ClearGrasp captures; the ViT-g variant is
    reported separately in the last row.}
    \label{tab:dc_sensor}
    \resizebox{\columnwidth}{!}{
    \begin{tabular}{l|cccccccc}
    \toprule
 & \multicolumn{3}{c|}{\bf{Hammer}}&\multicolumn{2}{c|}{\bf{Cleargrasp}}&\multicolumn{3}{c}{\bf{Lingbot}} \\
& D435 & L515 & Tof & D415 & D435 & D415 & D435 & D455\\
\midrule
CDMs~\cite{cdms} & {\bf 0.032}/{\bf 0.931} & \underline{0.032}/0.884 & 0.059/0.747 & \underline{0.017}/\underline{0.893} & \underline{0.013}/\underline{0.948} & 0.516/0.831 & 0.585/0.883 & 0.623/0.755 \\
OMNI-DC~\cite{omnidc} & 0.079/0.672 & 0.076/0.752 & 0.284/0.470 & 0.042/0.454 & 0.027/0.705 & 0.490/0.840 & 0.577/0.821 & 0.918/0.789 \\
Any2Full~\cite{AnyTF} & 0.050/0.775 & 0.066/0.841 & 0.066/0.728 & 0.030/0.673 & 0.024/0.751 & 0.611/0.902 & 0.633/0.903 & 1.450/0.813 \\
PriorDA~\cite{priorda} & 0.077/0.492 & 0.086/0.543 & 0.154/0.395 & 0.040/0.417 & 0.036/0.557 & 0.492/0.626 & 0.598/0.593 & 0.806/0.587 \\
LingBot-Depth 1.0~\cite{lingbot-depth} & 0.036/0.861 & 0.034/\underline{0.908} & \underline{0.035}/\underline{0.803} & 0.022/0.787 & 0.018/0.831 & \underline{0.373}/\underline{0.934} & {\bf 0.345}/\underline{0.942} & \underline{0.434}/\underline{0.877} \\
LingBot-Depth 2.0 & \underline{0.032}/\underline{0.895} & {\bf 0.023}/{\bf 0.942} & {\bf 0.031}/{\bf 0.912} & {\bf 0.010}/{\bf 0.981} & {\bf 0.012}/{\bf 0.960} & {\bf 0.242}/{\bf 0.949} & \underline{0.363}/{\bf 0.964} & {\bf 0.394}/{\bf 0.930} \\
\midrule
LingBot-Depth 2.0(ViT-g) & 0.031/0.916 & 0.022/0.950 & 0.025/0.952 & 0.016/0.897 & 0.014/0.917 & 0.228/0.957 & 0.345/0.968 & 0.375/0.940 \\
    \bottomrule 
    \end{tabular}
    }
\end{table}

\begin{table}[]
    \centering
    \caption{\textbf{Depth completion on block-masked and sparse input
    depth.} \normalfont In the block-mask regime, large contiguous
    regions of the input depth are removed; in the sparse regime, only
    scattered depth points are kept. Each cell reports
    RMSE$\downarrow$\,/\,$D_{105}\uparrow$; the best result per column
    is in bold and the second best is underlined. LingBot-Depth 2.0
    attains the best RMSE on seven of the eight benchmarks; the ViT-g
    variant is reported separately in the last row.}
    \label{tab:dc_block_sparse}
    \resizebox{\columnwidth}{!}{
    \begin{tabular}{l|cccccccc}
    \toprule
     & \multicolumn{4}{c|}{\bf{Block Mask}}&\multicolumn{4}{c}{\bf{Sparse}} \\
                  & DIODE-In & DIODE-Out & Ibims-1 & NYU & VOID & Ibims-1 & NYU & ETH3D \\
\midrule
CDMs~\cite{cdms} & 0.461/0.762 & 4.594/0.594 & 0.285/0.850 & 0.280/0.782 & 0.990/0.181 & 1.504/0.129 & 0.728/0.281 & 4.173/0.061 \\
OMNI-DC~\cite{omnidc} & 0.183/0.910 & 3.679/0.492 & 0.195/0.871 & 0.151/0.861 & 0.206/\underline{0.937} & \underline{0.132}/{\bf 0.973} & \underline{0.118}/\underline{0.949} & 0.600/0.872 \\
Any2Full~\cite{AnyTF} & \underline{0.104}/\underline{0.989} & 6.803/0.807 & \underline{0.148}/\underline{0.963} & 0.129/0.948 & 0.224/0.909 & 0.148/0.962 & 0.144/0.938 & 2.033/0.791 \\
PriorDA~\cite{priorda} & 0.307/0.713 & \underline{3.315}/0.596 & 0.356/0.653 & 0.352/0.543 & 0.211/0.912 & 0.138/0.960 & 0.126/0.936 & 0.562/0.840 \\
LingBot-Depth 1.0~\cite{lingbot-depth} & 0.132/0.981 & 3.404/\underline{0.809} & 0.188/0.955 & \underline{0.117}/{\bf 0.953} & \underline{0.199}/0.897 & 0.155/0.965 & 0.143/0.937 & {\bf 0.368}/{\bf 0.929} \\
LingBot-Depth 2.0 & {\bf 0.062}/{\bf 0.995} & {\bf 2.440}/{\bf 0.826} & {\bf 0.138}/{\bf 0.965} & {\bf 0.116}/\underline{0.951} & {\bf 0.170}/{\bf 0.940} & {\bf 0.131}/\underline{0.969} & {\bf 0.113}/{\bf 0.950} & \underline{0.522}/\underline{0.907} \\
\midrule
LingBot-Depth 2.0(ViT-g) & 0.060/0.996 & 2.298/0.841 & 0.132/0.969 & 0.114/0.954 & 0.174/0.940 & 0.124/0.973 & 0.113/0.951 & 0.476/0.924 \\
    \bottomrule 
    \end{tabular}
    }
\end{table}

\begin{figure}
    \centering
    \includegraphics[width=\linewidth]{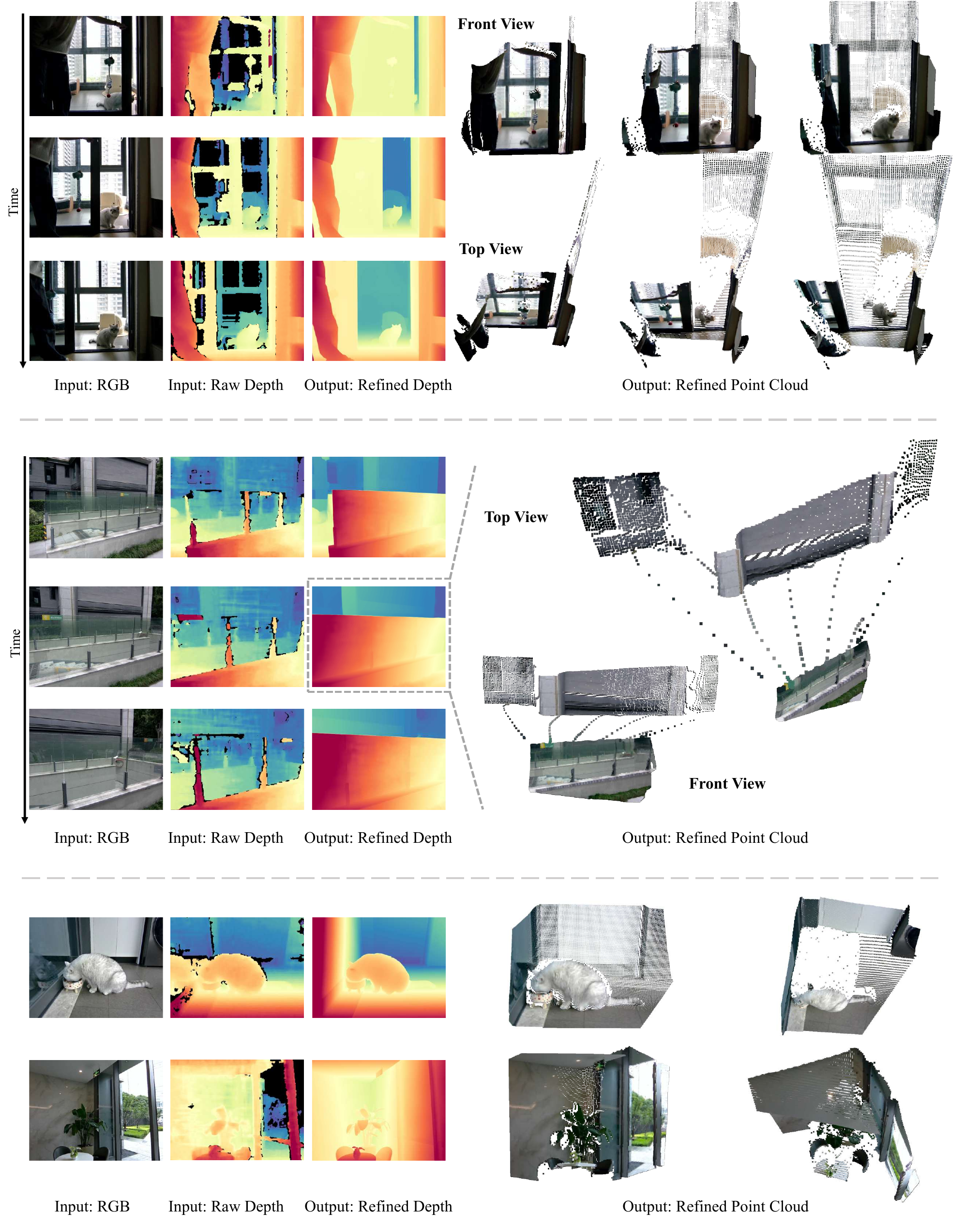}
    \caption{\textbf{LingBot-Depth 2.0 on mirror and glass scenes.}
    \normalfont Each group shows the input RGB, the raw sensor depth
    and the refined depth for consecutive frames of a captured sequence,
    together with front and top views of the refined point cloud. The
    raw depth is missing exactly on the hardest surfaces: window panes,
    a glass balustrade and reflective floors return no measurements.
    The completed regions form flat, contiguous planes in the point
    clouds and remain stable over time.}
    \label{fig:depth1}
\end{figure}

\begin{figure}
    \centering
    \includegraphics[width=\linewidth]{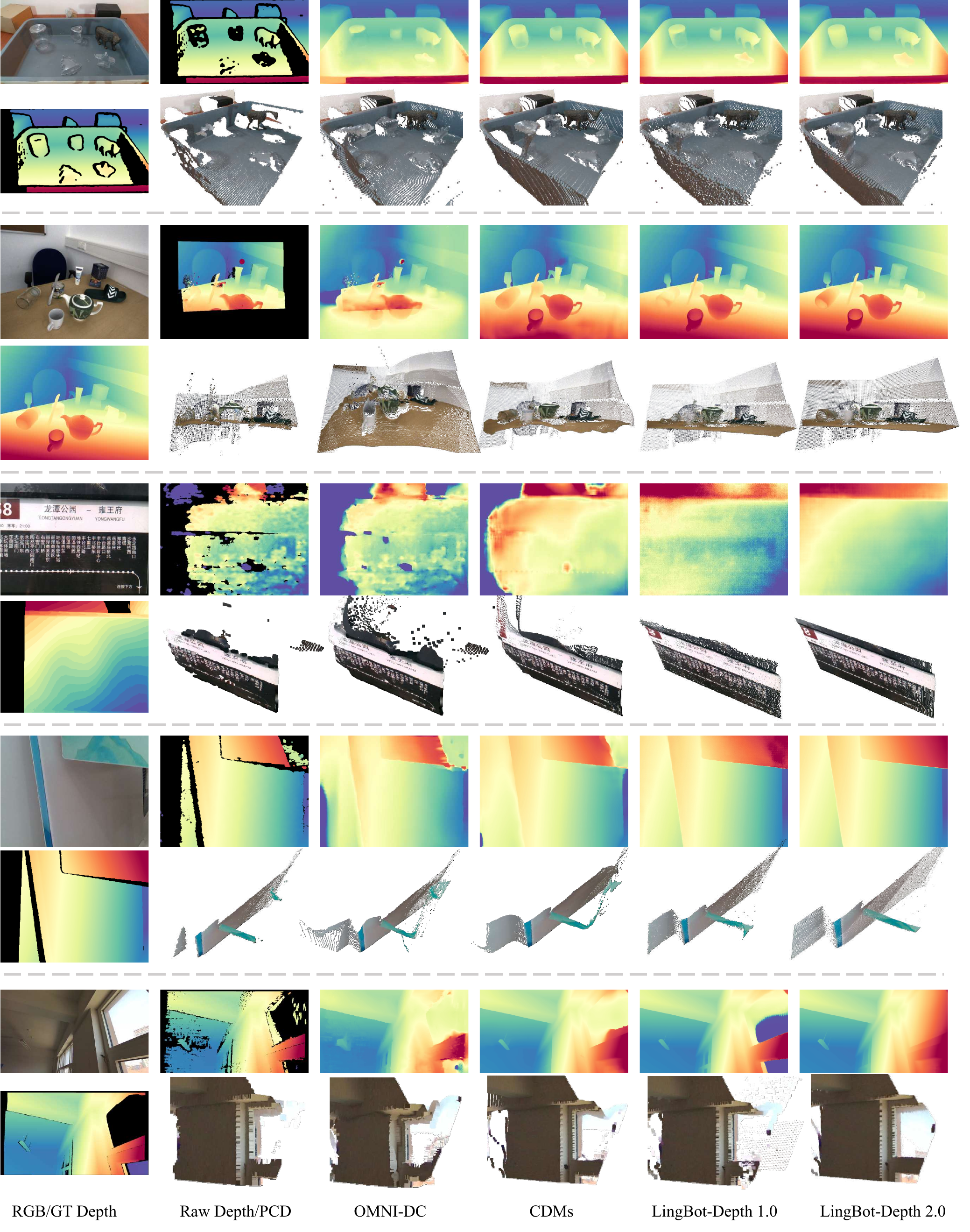}
    \caption{\textbf{Qualitative comparison of depth completion
    methods.} \normalfont Each scene shows the completed depth map
    (top) and the corresponding point cloud (bottom) for OMNI-DC, CDMs,
    LingBot-Depth 1.0 and LingBot-Depth 2.0, next to the RGB frame with
    ground-truth depth and the raw sensor input. The baselines leave
    missing regions open, bend planar surfaces, or scatter floating
    points near depth discontinuities; LingBot-Depth 2.0 keeps planes
    straight and object boundaries sharp.}
    \label{fig:depth2}
\end{figure}

\subsection{A Recap of LingBot-Depth 1.0}
\label{sec:depth:recap}

\paragraph{Masked depth modeling.} LingBot-Depth formulates depth
completion as masked depth modeling (MDM), an RGB-D variant of masked autoencoding~\cite{mae}. The key observation is that the missing measurements of commodity depth cameras are not nuisance noise but a natural masking signal: they concentrate exactly where imaging is hard, on specular, transparent and textureless surfaces, and thus mark the regions where geometry must be inferred from visual context. Accordingly, the RGB frame and the raw depth map are patchified by two separate embedding layers onto a shared token grid, and each token receives a shared spatial positional embedding plus a modality embedding that distinguishes the two sources. Depth tokens are masked according to sensor validity: fully invalid patches are always masked, partially valid patches are masked with high probability, and fully valid tokens are randomly added until the target masking ratio of 60\% to 90\% is reached. A ViT-L encoder, initialized from DINOv2~\cite{oquab2023dinov2}, jointly embeds the full set of RGB tokens and the unmasked depth tokens; the decoder discards the latent depth tokens and reconstructs the full-resolution depth map from the contextual tokens alone with a hierarchical ConvStack head adapted from MoGe~\cite{moge}, supervised by an L1 loss on valid ground-truth pixels. 

\paragraph{Data curation.} MDM requires RGB-D data whose missing patterns are realistic, whereas existing datasets either avoid hard imaging conditions or provide rendered, artifact-free depth. LingBot-Depth therefore curates its own data through two parallel pipelines. The synthetic stream renders RGB, perfect depth and speckle-projected stereo pairs in Blender from self-hosted indoor scenes, then runs semi-global matching on the stereo pair so that the resulting sensor-like depth carries the artifacts of real active cameras (1M samples from 442 scenes). The real-world stream builds on a custom multi-camera capture rig spanning diverse indoor, commercial and outdoor scenes, with pseudo ground truth distilled from the stereo infrared pairs by FoundationStereo~\cite{foundationstereo} under left-right consistency checks (2M captures). Combined with open-source RGB-D datasets, this yields roughly 10M training samples, of which the curated 3M were publicly released. Trained on this corpus, LingBot-Depth surpasses top-tier RGB-D cameras in both depth precision and pixel coverage.

\subsection{Two Important Tweaks for LingBot-Depth}
\label{sec:depth:tweaks}
LingBot-Depth 2.0 keeps the masked depth modeling recipe of
Sec.~\ref{sec:depth:recap} unchanged and improves its two external ingredients: the encoder initialization and the scale of the curated training data.

\paragraph{Spatial-perception-native encoder.} The single external dependency of the recipe is the encoder initialization, and that is precisely where LingBot-Depth 2.0 upgrades: the DINOv2 ViT-L/14 encoder is replaced by our \method pretraining, at both ViT-L/16 and ViT-g/16.  Table~\ref{tab:dc_encoder_init} isolates this choice by training the same MDM pipeline from different initializations. At ViT-L, the \method encoder improves over both the DINOv2 and DINOv3 initializations on nearly every benchmark, with the largest gains on the hardest block-mask patterns ({\em e.g.,} 0.094 vs. 0.152 RMSE on DIODE-Indoor against DINOv2), and the advantage carries to giant scale (0.083 vs. 0.118 on DIODE-Indoor).  This is the fit one would expect: depth completion concentrates its uncertainty around object boundaries and material transitions, exactly the structure that the boundary-anchored features of \method encode.

\paragraph{Scaling the curated data.} The second tweak scales the data: using the capture and pseudo-labeling machinery recapped in Sec.~\ref{sec:depth:recap}, the curated corpus grows from the publicly released 3M samples to 150M RGB-D samples.
Fig.~\ref{fig:data_size} tracks depth completion quality as the training set grows: accuracy improves monotonically with data for every encoder initialization compared, and the \method initialization retains its lead at every data scale. The two tweaks therefore compound: the better starting point does not get washed out by more data, it keeps paying at 150M.

\subsection{Benchmarking LingBot-Depth 2.0}
\label{sec:depth:results}

\paragraph{Datasets.} Our evaluation data falls into three categories, matching the structure of Tables~\ref{tab:dc_sensor} and~\ref{tab:dc_block_sparse}. (1)~\emph{Block-mask benchmarks} are public RGB-D datasets with high-quality ground truth, from which we remove large contiguous regions of the input depth: DIODE-Indoor and DIODE-Outdoor~\cite{diode}, precise laser scans covering both indoor and outdoor scenes; iBims-1~\cite{ibims1}, a laser-scanned indoor benchmark with carefully verified ground truth, particularly around depth boundaries; and NYU~\cite{nyuv2}, the standard Kinect-captured indoor test split. (2)~\emph{Sparse benchmarks} reduce the input depth to scattered measurements, sampled following MarigoldDC~\cite{MarigoldDC} on VOID~\cite{void}, iBims-1 and NYU, and taken from sparse SfM depth on ETH3D~\cite{eth3d}. (3)~\emph{Real-sensor benchmarks} evaluate on raw, incomplete captures of commodity depth cameras: HAMMER~\cite{hammer}, which pairs RealSense D435, L515 and time-of-flight captures of the same scenes with high-precision ground truth; ClearGrasp~\cite{cleargrasp}, real D415 and D435 captures of transparent objects, a worst case for active depth sensing; and our self-captured LingBot evaluation set with D415, D435 and D455 cameras.
Our LingBot evaluation set comprises $1{,}751$ frames captured across $35$ distinct indoor scenes. The scenes span a broad range of everyday environments: offices, meeting rooms and reception areas; schools, libraries and laboratories; hospitals and clinics; hotels, guestrooms and homes; restaurants, showrooms and retail spaces; and transit areas such as stations, parking lots, garages and elevators. Many of these contain the reflective, transparent and textureless surfaces that are a worst case for active depth sensing. To obtain reliable ground truth, we run a strong stereo-matching model (FoundationStereo~\cite{foundationstereo}) on the left--right image pairs of each camera; a left--right consistency check discards unreliable estimates, and a vision--language model (Qwen3.5~\cite{qwen}) is used to detect and remove frames with large motion blur. Pairing this ground truth with the raw, incomplete depth of each RealSense camera lets us evaluate depth completion under realistic sensing conditions.

\paragraph{Evaluation protocol.} We evaluate depth completion in two input regimes on the public benchmarks above, plus the raw captures of the real-sensor suite. In the block-mask regime, large contiguous regions of the input depth are missing; in the sparse regime, only scattered depth points are given. We report RMSE and the threshold accuracy $D_{105}$, and compare against recent depth completion systems (CDMs~\cite{cdms}, OMNI-DC~\cite{omnidc}, Any2Full~\cite{AnyTF}, PriorDA~\cite{priorda}) and LingBot-Depth 1.0.

\paragraph{Quantitative results.} On the masking patterns (Table~\ref{tab:dc_block_sparse}), LingBot-Depth 2.0 with the ViT-L backbone attains the best RMSE on seven of the eight benchmarks: all four block-mask settings and three of the four sparse settings, with ETH3D remaining with LingBot-Depth 1.0. The step over the previous version is large where completion is hardest: on block-masked DIODE-Indoor the RMSE is halved (0.132 to 0.062), and DIODE-Outdoor improves from 3.404 to 2.440. The external baselines are strong only within their preferred regime: CDMs collapses under sparse inputs and Any2Full degrades on outdoor block masking, whereas LingBot-Depth 2.0 is consistent across both. On real sensors (Table~\ref{tab:dc_sensor}), 2.0 is the best on six of the eight camera configurations, matches CDMs on Hammer D435, and is particularly strong on the transparent-object ClearGrasp captures (0.010 and 0.012 RMSE), the classic failure case of active depth sensing. Scaling the backbone to ViT-g pushes further on most pattern benchmarks (0.060 on DIODE-Indoor, 0.114 and 0.113 on NYU in the two regimes) and on the LingBot sensor suite (0.228, 0.345 and 0.375 across the three cameras), with ClearGrasp remaining in favor of the ViT-L variant. 

\paragraph{Qualitative results.}
Fig.~\ref{fig:depth1} probes the two classic failure modes of active depth sensing, mirrors and glass, on captured video sequences. The raw sensor depth is missing exactly where the scenes are hardest: window panes, a glass balustrade and reflective floors return no measurements at all. LingBot-Depth 2.0 completes these regions with geometry that is consistent as a surface, not merely plausible per pixel: in the refined point clouds, the window and the balustrade appear as flat, contiguous planes in both the front and top views, and the completion stays stable across consecutive frames of each sequence. Fig.~\ref{fig:depth2} compares against the baselines on transparent tabletop objects, outdoor signage, dark walls with thin structures, and a bright window scene.  The differences are most visible in the point-cloud rows: the baselines either leave the missing regions open, bend large planar surfaces, or scatter floating points around depth discontinuities, and LingBot-Depth 1.0 still exhibits residual outliers near windows and object rims. LingBot-Depth 2.0 keeps walls straight, planes intact and object boundaries sharp, with occlusion edges in the completed depth that align with the object contours of the RGB frame, the signature of its boundary-aware encoder.

\section{Conclusion}
\label{sec:conclusion}

This report presented \method, a vision foundation model pretrained to be
spatial-perception native. The central idea is to make boundary structure
a first-class citizen of self-supervised learning: image boundaries are
represented as dense categorical boundary fields, boundary-bearing tokens
are forced into the masking pattern, and the training targets are
bootstrapped online from the teacher's own predictions, validated by a
parameter-free a-contrario test whose null hypothesis coincides with the
uniform categorical distribution. Trained at ViT-g scale on curated data,
\method delivers the strongest frozen-feature dense prediction in its
class, including the best NYUv2 linear-probing RMSE of all compared
models, ahead of the 7B DINOv3, video object segmentation on par with the
strongest distilled models, competitive image-level recognition, and
boundary tokens stable enough to be tracked through video. Distillation
carries these properties to ViT-L, ViT-B and ViT-S students, so the same
recipe serves every deployment budget.

The value of a vision foundation model is ultimately measured downstream.
Upgrading \taskold to \tasknew changed nothing in the masked depth
modeling recipe except the encoder initialization and the scale of the
curated data, and delivered the leading depth completion performance
across masking patterns and real depth cameras, with the advantage of the
\method initialization widening as the training data grows. We take this
as evidence that boundary-oriented pretraining is a practical route
toward robotic spatial perception, and we release the pretrained models
of \method to the community.

\section*{Acknowledgments}
We gratefully acknowledge the optical testing team at Orbbec Inc. for their professional support.
We also thank Yubin Hu, Changkun Liu, Xingyi He, Mengfei Xia, Jianyuan Wang, Shangzhan Zhang, Yuanbo Yang for helpful discussion.
The large-scale curated RGB-D data for LingBot-Depth 2.0 cannot be successful without the support of Yongtao Huang, Yibo Lu, Chongjun Zhong, Han Zhang and Xiage Qin.

\justifying

\section*{Appendix}
\appendix
\section{Sampling Boundary Fields}
\label{app:sampling}
This appendix details the demonstration behind Finding~\ref{finding:sampling} and Fig.~\ref{fig:sampling}: decoding line segments from boundary fields that carry no learned information.

\paragraph{Setup.} Images are resized to $512\times512$ and the boundary field is defined at output stride 2, with every channel normalized to $[0,1]$: the distance channel by the support threshold $\tau_d$ (5\,px at the field resolution) and the three angular channels by their angular ranges (Sec.~\ref{sec:method:categorical}). Corner points are produced by the frozen corner-point detector used throughout pretraining (Sec.~\ref{sec:method:bootstrap}) and are kept fixed across all draws.

\paragraph{Sampling.} For each draw, the field channels are sampled i.i.d.\ per position from $\mathcal{U}(0,1)$, with the sub-pixel refinement offsets set to zero. In the bottom rows of Fig.~\ref{fig:sampling}, \emph{all} channels are random, including the direction channel $\theta$ that orients each position's chord proposal; randomly oriented proposals rarely agree, so the decoded segments stay short, yet they remain anchored on the corner points. In the top rows, only $\theta$ is replaced by a parameter-free guidance that carries no learned information: the image's \emph{level-line} orientation, the direction orthogonal to the local intensity gradient that underlies classical a-contrario line detection~\citep{lsd,desolneux}. Collinear level-lines are grouped into line-support regions~\citep{lsd}, the resulting segments are rendered into a boundary field (the lifting of Sec.~\ref{sec:method:categorical}), and the $\theta$ channel of this rendering replaces the random draw; the distance and endpoint channels stay random.

\paragraph{Decoding.} Decoding follows Sec.~\ref{sec:method:bootstrap} with the a-contrario validation disabled: every position proposes a chord from its attributes $(d, \theta, \phi^{1}, \phi^{2})$, the proposal endpoints are attached to their nearest corner points, votes are accumulated over every corner pair, and the pairs with sufficient votes become the decoded segments. The five columns of Fig.~\ref{fig:sampling} are independent draws, rendered while sweeping the minimum vote count from 2 to 10; the decoded line segments are essentially unchanged across both the draws and the sweep. The randomness in $d$, $\phi^{1}$ and $\phi^{2}$ is thus fully absorbed by corner anchoring and vote aggregation, while the direction channel controls the completeness of the recovery: the fully random field still anchors on the corners but yields only short fragments (bottom rows), and the parameter-free level-line guidance recovers the complete line segments (top rows).

\section{A-contrario Validation}
\label{app:nfa}
\begin{figure}[t]
\centering
\setlength{\tabcolsep}{1pt}
\renewcommand{\arraystretch}{0}
\begin{tabular}{cccc}
{\scriptsize\shortstack{(a) detector corners\\w/o validation}} &
{\scriptsize\shortstack{(b) detector corners\\w/ validation}} &
{\scriptsize\shortstack{(c) field-derived corners\\w/o validation}} &
{\scriptsize\shortstack{(d) field-derived corners\\w/ validation}} \\
\noalign{\vskip 2pt}
\includegraphics[width=0.243\linewidth]{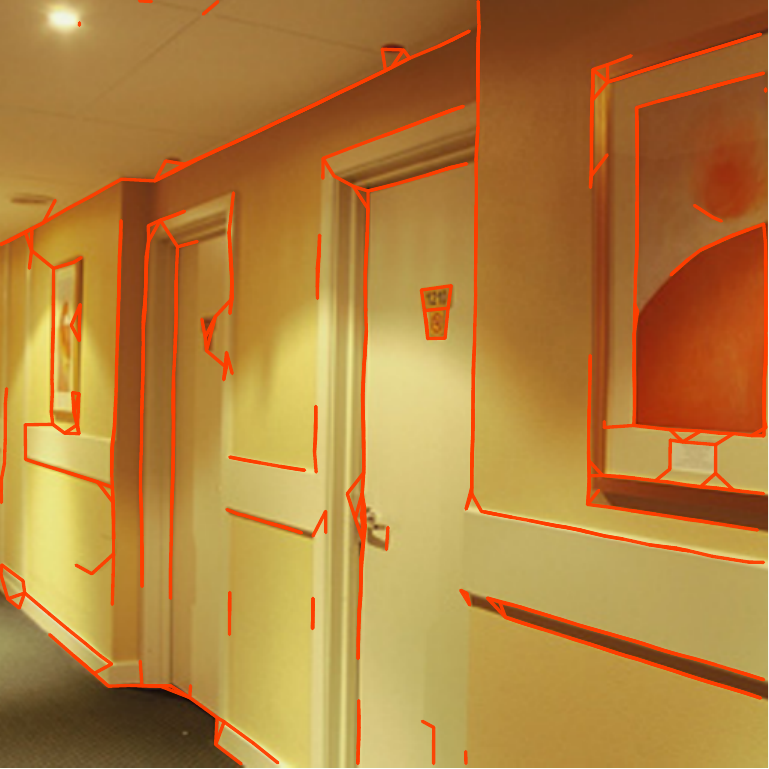} &
\includegraphics[width=0.243\linewidth]{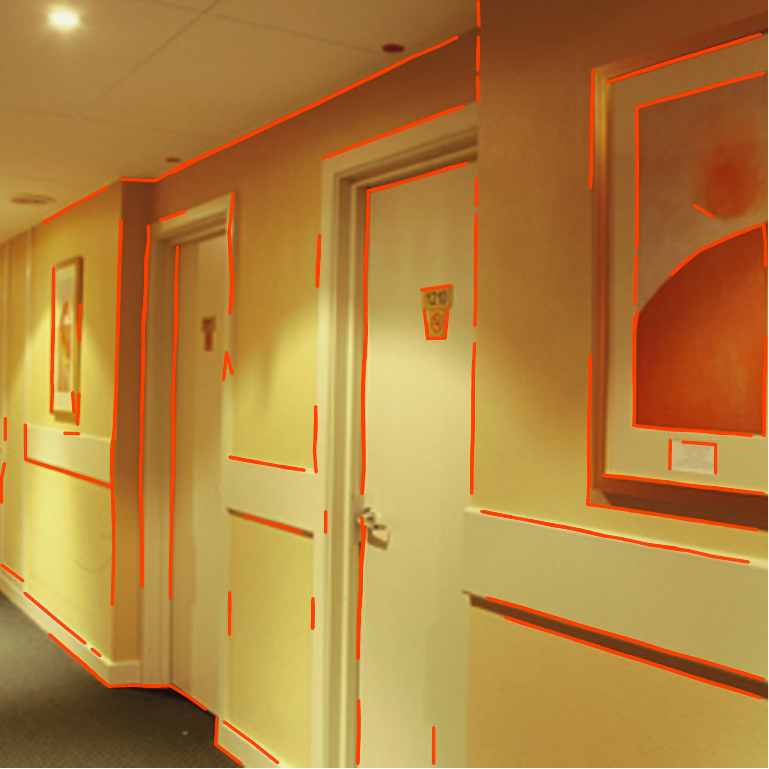} &
\includegraphics[width=0.243\linewidth]{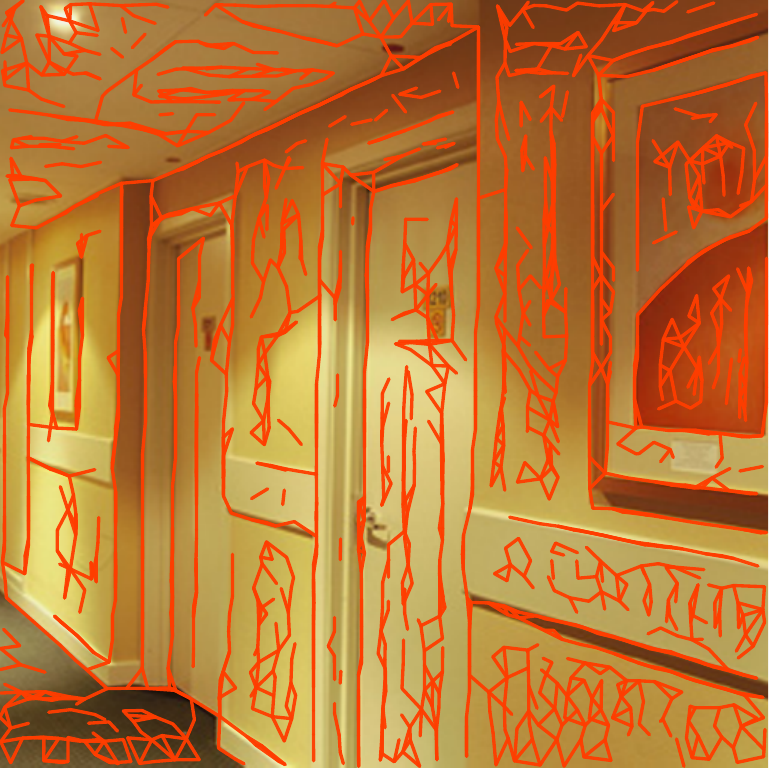} &
\includegraphics[width=0.243\linewidth]{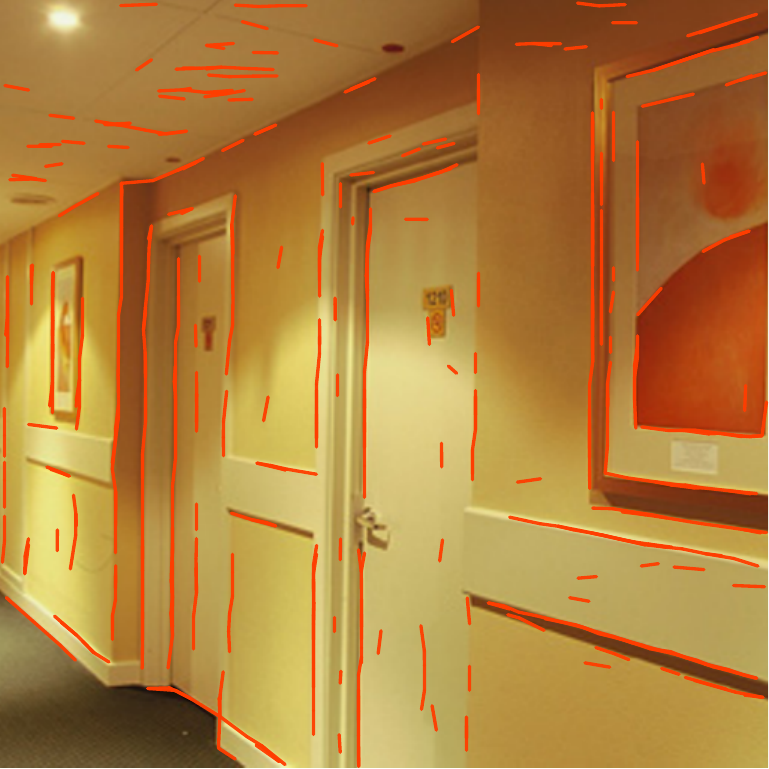} \\
\end{tabular}
\caption{\textbf{Corner anchors and a-contrario validation are complementary safeguards.}
\normalfont The same boundary field, decoded from the same \method teacher under identical
settings, with corner points taken either from the frozen single-block detector
(a, b; 920 candidate segments) or derived from the field itself via
$\mathrm{jloc} = 1 - d$ (c, d; 2,382 candidates). With reliable corner anchors, the
vote-aggregation decoding is already clean before any validation (a), and the
a-contrario test removes only minor spurious residues (b; 673 segments kept). With
weak, self-derived anchors, the unvalidated decoding degenerates into dense spurious
structure (c), yet the a-contrario test recovers almost identical line segments
(d; 675 kept, 72\% rejected). Either safeguard alone suffices; only when both are
removed does decoding fail. Together with Finding~\ref{finding:sampling}, this
explains why boundary targets can be bootstrapped robustly from scratch.}
\label{fig:anchor_nfa}
\end{figure}

The a-contrario framework~\citep{lsd,desolneux} formalizes the Helmholtz principle: a geometric structure is meaningful only if it would be highly unlikely under a null model that contains no structure. For a candidate segment with $n$ support pixels, of which $k$ have orientations aligned with the segment within a tolerance, the Number of False Alarms (NFA) is
\begin{equation}
    \mathrm{NFA}(n,k) = N_{\mathrm{t}} \cdot \sum_{i=k}^{n}\binom{n}{i}\, p^{\,i}\,(1-p)^{\,n-i},
\end{equation}
where $p$ is the probability that a pixel aligns with the segment by chance and $N_{\mathrm{t}}$ is the number of tested candidates. A segment is accepted iff $\mathrm{NFA} \le \varepsilon$; the canonical choice $\varepsilon = 1$ (at most one false alarm per image on average) makes the test essentially parameter-free. The null hypothesis, uniformly distributed orientations, is exactly the uniform distribution over bins of our categorical boundary field (Sec.~\ref{sec:method:categorical}).

\paragraph{Why the classical test does not parallelize.}
In LSD~\citep{lsd}, candidate formation and validation are entangled. A candidate rectangle only exists after a greedy region-growing pass: starting from a high-gradient seed pixel, the region absorbs neighbors whose level-line orientation agrees within tolerance and marks them as consumed, so that later seeds cannot reuse them; a rectangle is then fitted to the grown region and submitted to the NFA test. Regions are grown one at a time, the outcome depends on the seed order, and the control flow branches on every pixel, which is why a-contrario detectors have remained sequential CPU designs.

\paragraph{Corner anchoring decouples candidates from validation.}
Our pipeline obtains the candidate set without any grouping. Because the boundary field is deliberately redundant, a complete chord proposal, with both endpoints, can be read out densely at every position (Sec.~\ref{sec:method:categorical}). Corner points then anchor these proposals: a batched line-segment matcher snaps the two endpoints of every proposal to their nearest corner points, with the pairwise distance computation fused into the argmin reduction so that memory stays linear in the number of proposals, and each corner pair accumulates the votes of the proposals assigned to it. The candidate set is simply the resulting corner-pair graph. Validation thereby becomes a pure per-candidate reduction with no shared state: one GPU thread per candidate counts the support and aligned pixels in a fixed-width rectangle around its segment and evaluates the binomial tail above, so all candidate segments of a training batch are validated in parallel (Sec.~\ref{sec:scaling:infra}). This decoupling, dense endpoint readout plus corner anchoring in place of region growing, is what makes a-contrario validation cheap enough to run inside the training loop at every iteration.

\paragraph{Implementation constants.}
The orientation evidence is the image's level-line field, computed with LSD's $2\times2$ finite-difference gradient on the grayscale view, batched over the global crops; pixels with negligible gradient magnitude are marked as undefined and never count as aligned. Each candidate uses a rectangle of fixed width $3$\,px, and orientations are compared modulo $\pi$ with an alignment tolerance of $\pi/16$, i.e., $p = 1/16$, stricter than the $\pi/8$ of LSD. We keep the classical number of tests $N_{\mathrm{t}} = (HW)^{5/2}$, an upper bound over all possible rectangles; this is conservative for our candidate set, which contains only the corner pairs, and we drop the extra factor LSD adds for its width refinements since our width is fixed. The acceptance threshold is the canonical $\varepsilon = 1$, and candidates whose aligned-pixel density falls below $0.5$ are rejected before the tail is evaluated. During pretraining the test runs at every iteration on every decoded segment, and images retaining fewer than $10$ validated line segments are excluded from the boundary loss for that iteration. The same reduction applies unchanged when the orientation evidence is taken from the model's own boundary field instead of the image gradient, which is how the boundary tokens of Fig.~\ref{fig:teaser} are validated.

{\small
\bibliographystyle{plain}
\bibliography{references}
}
\end{document}